\documentclass{article} 
\usepackage{iclr2025_conference,times}

\usepackage{hyperref}
\usepackage{url}
\usepackage{graphicx} 
\usepackage{amsmath}
\usepackage{amssymb}
\usepackage{booktabs}
\usepackage{multicol,multirow}

\title{Boosting Sclera Segmentation through Semi-supervised Learning with Fewer Labels}

\author{Guanjun Wang, Lu Wang, Ning Niu, Qiaoyi Yao, Yixuan Wang, Sufen Ren\thanks{Corresponding Author},  \\
Hainan University\\
\texttt{Aug\_0815@163.com} \\
\AND
Shengchao Chen\thanks{Corresponding Author} \\
University of Technology Sydney \\
\texttt{shengchao.chen.uts@gmail.com}
}

\iclrfinalcopy 
\begin{document}

\maketitle

\begin{abstract}
Sclera segmentation is crucial for developing automatic eye-related medical computer-aided diagnostic systems, as well as for personal identification and verification, because the sclera contains distinct personal features. Deep learning-based sclera segmentation has achieved significant success compared to traditional methods that rely on hand-crafted features, primarily because it can autonomously extract critical output-related features without the need to consider potential physical constraints. However, achieving accurate sclera segmentation using these methods is challenging due to the scarcity of high-quality, fully labeled datasets, which depend on costly, labor-intensive medical acquisition and expertise. To address this challenge, this paper introduces a novel sclera segmentation framework that excels with limited labeled samples. Specifically, we employ a semi-supervised learning method that integrates domain-specific improvements and image-based spatial transformations to enhance segmentation performance. Additionally, we have developed a real-world eye diagnosis dataset to enrich the evaluation process. Extensive experiments on our dataset and two additional public datasets demonstrate the effectiveness and superiority of our proposed method, especially with significantly fewer labeled samples.
\end{abstract}

The sclera plays a pivotal role in various fields, including computer-aided diagnostic (CAD) systems~\citep{chen2023interpretable,ren2024federated,peng2023diffusion,wang2024ensemble,shi2023recognition} and personal identification~\citep{alshakree2024human}. In the context of CAD, changes in the sclera serve as critical indicators for diagnosing and monitoring ocular diseases such as dry eyes, glaucoma, and cataracts~\citep{acharya2015automated,noori2016scleral,mookiah2012data}. This enables physicians to conduct comprehensive examinations of ocular structures, identify abnormalities, and assess treatment outcomes effectively. Furthermore, scleral segmentation offers a reliable and secure method for personal identification and verification in biometrics, leveraging the unique vascular patterns of the sclera found in each individual~\citep{zhao2017towards,27}.

Achieving robust and reliable scleral recognition is crucial; however, it continues to face significant challenges in real-world applications. Effective scleral recognition primarily relies on precise vessel pattern extraction, which is contingent upon accurate scleral segmentation. This task is complicated by factors such as image reflections, eyelash occlusion, and suboptimal eyelid opening during eye image capture~\citep{29}. Moreover, accurate segmentation is vital to ensure that only relevant scleral regions containing vascular patterns are identified. Inaccurate segmentation can result in substantial information loss or the inclusion of irrelevant regions, thereby diminishing the effectiveness of the recognition system~\citep{73}. The challenge of achieving accurate scleral segmentation is particularly pronounced in uncontrolled environments, where factors such as blurring, occlusion, and varying illumination conditions can lead to further inaccuracies.

Recent advancements in sclera segmentation have shifted from labor-intensive, expensive hand-crafted feature extraction to advanced deep learning (DL)-based methods~\citep{chen2023foundation,chen2024free}. These methods can effectively build potential relationships between eye images and focused regions without considering the underlying physical relationships. For instance, Radu et al.~\citep{17} introduced a two-stage multi-classifier system for accurate patch-based segmentation. Rot et al.\citep{18, 35} proposed an Encoder-Decoder architecture~\citep{chen2023tempee} for comprehensive eye segmentation. Lucio et al.\citep{35} employed a cascaded model comprising a fully convolutional network (FCN) and a generative adversarial network (GAN) for full-image segmentation. Although these methods achieve excellent segmentation performance, their efficacy relies heavily on the availability of high-quality training datasets. However, acquiring such datasets is challenging, as medically relevant data typically involves high acquisition costs and complex labeling processes that require expert knowledge~\citep{fuhl2016pupilnet, swirski2012robust}. While techniques based on data augmentation and generative models~\citep{chen2022dynamic} can mitigate this issue to a certain extent, the generated or augmented images often lead to significant decision biases in the trained models. This is problematic in practice due to the low fault tolerance required in medical~\citep{ren2024federated} or security applications~\citep{chen2023prompt,ren2024distributed,chen2024personalized}.

To overcome the challenges and develop a robust, high-performance scleral segmentation framework, this paper proposes a Semi-Supervised Learning (SSL)-based method that reduces dependence on extensive labeled data in DL-based scleral segmentation. Specifically, SSL enhances model training by utilizing unlabeled data to improve learning from a limited number of labeled samples \citep{zhou2003learning}. The primary objective of SSL is to mitigate overfitting when using a small labeled dataset. Formally, given a dataset \(X = \{ x_1, x_2, \ldots, x_n \}\) with only the first \(k\) samples labeled as \(\{ y_1, y_2, \ldots, y_k \} \in Y\), and the rest unlabeled, the aim is to derive a function \(f : X \to Y\) that exploits the inherent relationships within the data to predict labels for the unlabeled samples. A prevalent inductive bias used to regulate SSL algorithms is the consistency assumption, which posits that predictions should remain stable even when data points or their representations are altered \citep{tarvainen2017mean, 68}. This principle has been effectively implemented in various frameworks by modifying either the input data \citep{68} or their latent representations \citep{gyawali2019semi} to reduce discrepancies in label predictions. While these methods have shown substantial improvements in generalization across different medical imaging tasks, their application in semantic segmentation is still limited. In segmentation, the output must maintain spatial coherence with the input pixel configuration, which complicates the application of consistency assumptions typical in classification tasks. As such, the use of SSL in semantic segmentation remains relatively unexplored, with only a few studies investigating its potential \citep{ouali2020semi}. This limitation is especially pronounced in eye segmentation applications, where the employment of SSL is particularly constrained \citep{32}. Our proposed method effectively addresses these issues in the context of scleral segmentation. This paper's contributions are threefold.
\begin{itemize}
    \item We propose a novel sclera segmentation framework utilizing a semi-supervised learning strategy that demonstrates exceptional performance, even with a limited number of labeled samples.
    \item We have developed a new real-world eye diagnosis dataset, comprising approximately 800 images from over 100 patients, each manually annotated under the guidance of medical experts, to enhance the evaluation process.
    \item Extensive testing on our dataset and two additional public datasets, namely UBIRIS.V2 and SBVPI, confirms the effectiveness and superiority of our proposed method, particularly when utilizing significantly fewer labeled samples.
\end{itemize} 

\section{Related Works}
\subsection{Hand-crafted Features-based Scleral Segmentation}
Hand-crafted is a key factor of traditional image processing methods~\citep{wang2024less,chen2023mask}. Scleral segmentation techniques that employ hand-crafted local features are typically divided into three categories: manual methods, pixel threshold-based methods, and shape contour-based methods. Manual segmentation approaches \citep{3,4}, while effective, are impractical for real-world applications due to their extensive processing time and the need for continuous human oversight. Pixel threshold methods \citep{5,6,7,8,9,10} perform well under controlled lighting conditions but falter in noisy or distorted scleral images. Shape contour-based methods \citep{11,12,13,14,15,16}, although useful in certain scenarios, face challenges such as the need to manually exclude occluded or noisy images. Moreover, the accuracy of these methods can be compromised by the sclera's borders, which may reduce the visibility of vascular areas or introduce artifacts from the eyelid or eyelashes, thereby degrading system performance. In contrast, advancements in deep learning have significantly improved image processing tasks \citep{17,18,19}, offering substantial enhancements over traditional hand-crafted feature-based methods.

\subsection{Deep Learning-based Scleral Segmentation}
Deep learning-based scleral segmentation can be categorized into image patch-based and full-image-based approaches. In the image patch-based category, Radu et al. \citep{17} introduced a two-stage multi-classifier system that analyzes 60 scleral and non-scleral regions from the UBIRIS.v1 dataset. Initially, a standard classifier is used, followed by a neural network classifier that operates within the generated probability space. In the full-image-based category, Rot et al. \citep{18,20} developed an encoder-decoder model, SegNet \citep{21}, for comprehensive eye segmentation. Lucio et al. \citep{19} proposed using a fully convolutional network (FCN) \citep{23} and a generative adversarial network (GAN) \citep{25}, where the FCN omits fully connected layers following the suggestions of Teichmann et al. \citep{24} with modifications to the VGG-16 architecture. Fadi et al. \citep{26} introduced two models for precise scleral segmentation: the multi-scale segmentation network (Eye-MS) and its more compact version, the small multi-scale segmentation network (Eye-MMS), which achieves similar performance with fewer parameters. Naqvi et al. \citep{27} developed Sclera-Net, a residual encoder-decoder network that utilizes identity and non-identity mappings for accurate delineation of scleral regions. Wang et al. \citep{28} improved the U-Net model to create ScleraSegNet, which incorporates an attention module to enhance feature discrimination. This method demonstrated notable accuracy on the UBIRIS.v2 and MICHE databases. In 2023, Wang et al. \citep{30} integrated meta-learning into the field, employing a meta-sampling strategy and UNet 3+ to facilitate cross-domain knowledge transfer. Sumanta et al. \citep{29} introduced a scleral recognition approach using a lightweight model, DSEG, based on UNET for efficient and accurate segmentation. Li et al. \citep{73} proposed Sclera-TransFuse, a two-stream hybrid model that combines ResNet and SwinTransformer for improved segmentation.

This paper employs a limited amount of labeled data supplemented by a substantial volume of unlabeled data, contrasting with methods that depend on extensive labeled datasets. Given the relative ease of acquiring unlabeled data and the challenges of collecting and labeling extensive datasets, semi-supervised learning (SSL) has become increasingly popular for real-world applications. This approach not only reduces data privacy risks but also prevents potential leaks of patient information, which are concerns in extensive dataset labeling.

\section{Preliminary}
\subsection{Sclera Segmentation}
Sclera segmentation aims to accurately identify the scleral region in ocular images, which can be denoted as a binary segmentation problem. Given an input image \( I \), the goal is to produce a binary mask \( M \) where:
\begin{equation}
M(x, y) = 
\begin{cases} 
1 \& \text{if pixel } (x, y) \text{ belongs to the sclera} \\
0 \& \text{otherwise}
\end{cases}
\end{equation}
The accuracy of segmentation can be quantified using metrics such as Intersection over Union (IoU):

\begin{equation}
    \text{IoU} = \frac{|M_{pred} \cap M_{true}|}{|M_{pred} \cup M_{true}|},
\end{equation}
where $M_{pred}$ is the predicted mask and $M_{true}$ is the ground truth mask. Challenges such as image reflections, eyelash occlusions, and suboptimal eyelid openings can be modeled as noise $N$ affecting the input image, leading to the observed image $I'$: $I' = I + N$. Effective segmentation methods must account for this noise to ensure that only relevant scleral regions are recognized.
\subsection{Semi-Supervised Learning}
Semi-supervised learning have been widely used in data-limited applications~\citep{peng2023diffusion,wang2022semi,ren2023unsupervised}. In semi-supervised learning, we denote the labeled dataset as $D_l = \{(x_i, y_i)\}_{i=1}^{N_l}$ and the unlabeled dataset as $D_u = \{x_j\}_{j=1}^{N_u}$. The objective is to minimize a combined loss function $L$ that incorporates both labeled and unlabeled data:
\begin{equation}
    L = L_l + \lambda L_u,
\end{equation}
where $L_l$ is the loss on labeled data, $L_u$ is the loss on unlabeled data, and $\lambda$ is a weighting parameter controlling the contribution of the unlabeled loss. The overall goal is to learn a function $f$ that predicts the segmentation mask $M$ by minimizing the expected loss:
\begin{equation}
    \mathbb{E}[L] = \mathbb{E}_{(x,y) \in D_l}[L_l(f(x), y)] + \mathbb{E}_{x \in D_u}[L_u(f(x))].
\end{equation}
This approach enables the model to leverage a large amount of unlabeled data to improve the segmentation accuracy, particularly in medical imaging applications where labeled data is scarce.
\section{Methodology}
\subsection{Overview}
\begin{figure*}[tbh]
	\centering
 	 \includegraphics[width=1\textwidth]{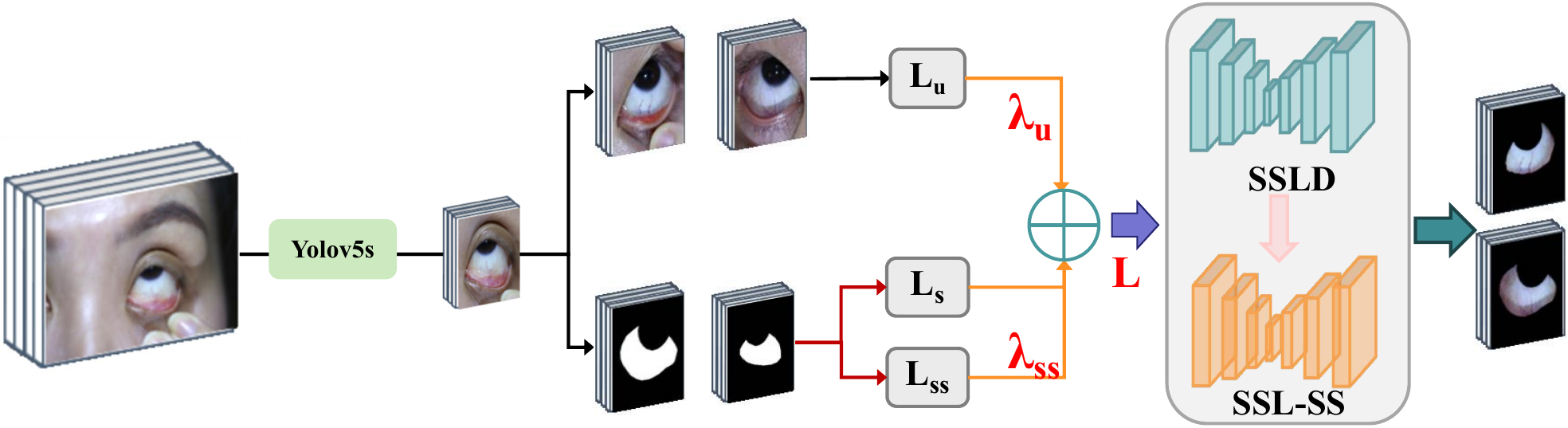}
 	  \caption{Overall structure of our method}\label{Figure.1}
 \end{figure*}

The overall framework of this study is illustrated in Fig.\ref{Figure.1}. In the data preprocessing phase, we employed a yolov5s-based automatic detection system\citep{74} to crop the eye region, significantly reducing background interference from areas outside the eyes and thereby minimizing subsequent challenges during scleral segmentation. This approach not only conserves significant time but also reduces the costs associated with manual cropping. Once the ocular dataset and corresponding labels were prepared, the semi-supervised learning (SSL) model was utilized to train the dataset, enabling the extraction of scleral images suitable for intelligent scleral recognition.

\subsection{Data Augmentation}
To overcome the challenge of insufficient data, data augmentation is the most common solution~\citep{xu2023dual,chen2022fabry,chen2022cmt,chen2022reconstruction}. We incorporated domain-specific and transformation-based augmentation techniques into our semi-supervised learning (SSL) framework. Specifically, domain-specific augmentation employs methods such as Contrast Limited Adaptive Histogram Equalization (CLAHE) \citep{pizer1987adaptive,37} and gamma correction, proven effective for enhancing eye segmentation tasks \citep{38}. Histogram equalization improves image contrast by redistributing intensity values to achieve a more uniform histogram distribution, thus enhancing image clarity. CLAHE, an advanced version of histogram equalization, limits noise amplification by capping contrast enhancements at specific pixel values. It adjusts the enhancement by modifying the slope of the transformation function based on the local cumulative distribution function (CDF) of pixel values. CLAHE sets a clipping threshold, typically between 3 and 4, to control excessive contrast enhancement. Instead of discarding surplus contrast, it redistributes this uniformly across the histogram, preserving overall brightness and contrast \citep{37}.

Gamma correction modifies the gamma curve of an image to adjust tones non-linearly, enhancing the contrast between darker and lighter areas and improving visual quality by highlighting details in both low and high intensity regions \citep{guan2009image}. By fine-tuning the gamma values, we selectively enhance image features crucial for effective segmentation. These augmentation techniques collectively enhance the model's robustness against variations in lighting and contrast, which are common challenges in eye image datasets. n the initial phase, domain-specific augmentation is applied to both labeled and unlabeled data. For labeled data, this augmentation exclusively uses CLAHE, employing identical random clip parameters and grid sizes as those used for unlabeled data. For each iteration of unlabeled data, we apply rotation and translation augmentations based on image transformation probabilities $p_1$ and $p_2$, respectively, denoted by \textit{T}.

\subsection{Semi-supervised Segmentation Model}
The semi-supervised segmentation method proposed in this paper is an improvement on the SSL framework proposed by Chaudhary et al.\citep{32}, which uses the SSL paradigm to process unlabeled dataset and improved U2Net as the segmentation network. 

\begin{figure*}[ht]
	\centering
 	 \includegraphics[width=\textwidth]{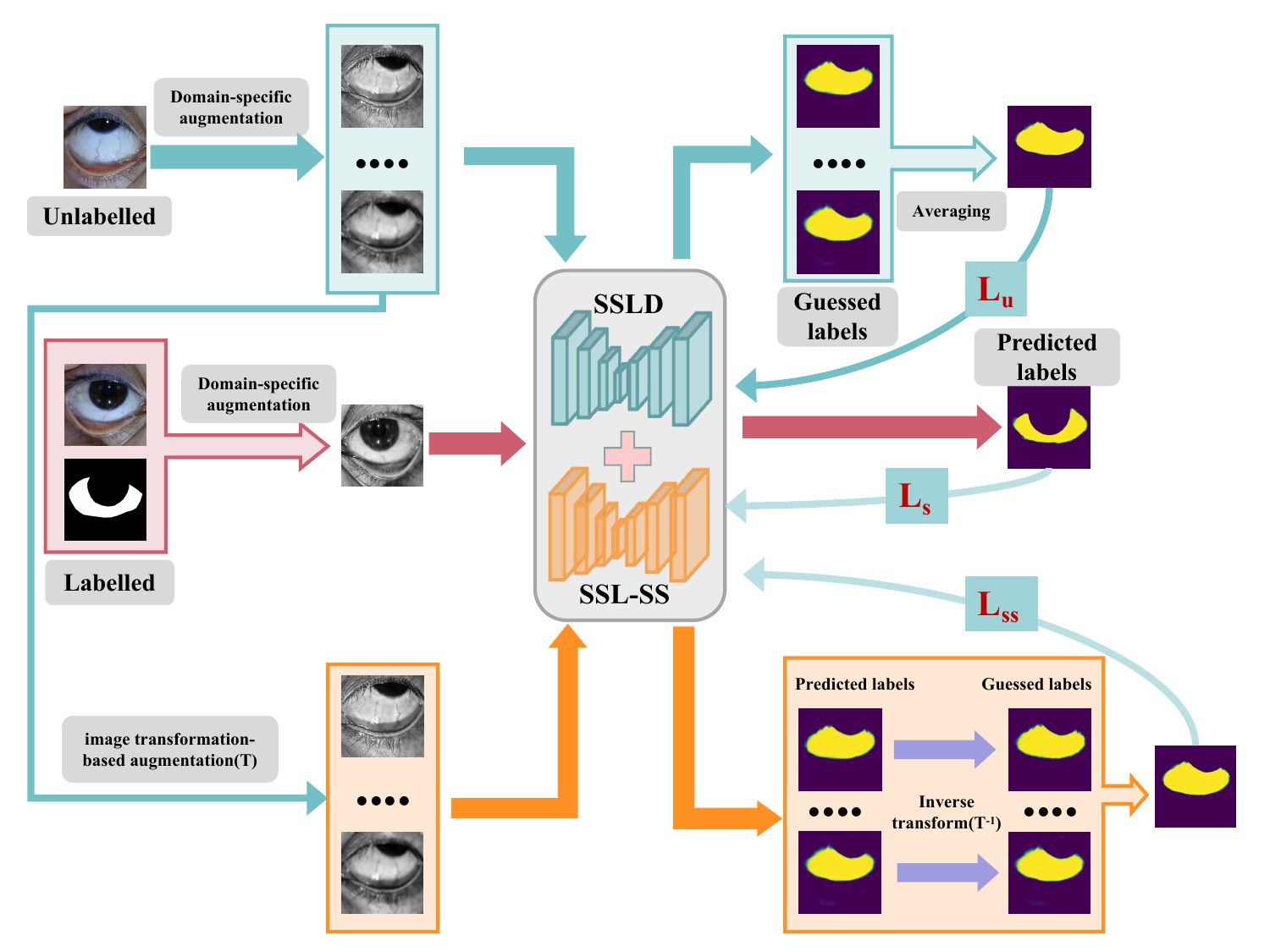}
 	  \caption{Structure of the SSL framework}\label{Figure.2}
 \end{figure*}

\subsubsection{SSL framework}
The SSL approach used, as shown in Fig.\ref{Figure.2}, is divided into two stages. The first stage involves SSL with domain-specific enhancements (SSLD), while the second stage utilizes SSL with self-supervised learning (SSL-SS). Compared to the first stage, the SSL-SS stage introduces a more effective regularization strategy by generating labels through a self-supervised mechanism.

In the dataset, there are \textit{n} labeled images $X_\textit{l}$ with corresponding labels $Y_\textit{l}$, and \textit{m} unlabeled images $X_\textit{u}$, where \textit{n} $<$ \textit{m}. For the segmentation task, the data space is represented as $X\in R^{C\times H\times W} $, where $H \times W$ denotes the spatial dimensions and $C$ represents the number of input channels. Similarly, the label space is denoted as $Y \in R^{P \times H \times W}$, where $P$ indicates the number of classes. Our objective is to learn the parameters $\theta$ of the mapping function \textit{f}: $x \rightarrow y$ through the segmentation network.

In the first stage (SSLD), domain-specific augmentations are applied, where each data point, both labeled and unlabeled, is augmented \textit{k} times, and these augmented copies are fed into the model to predict their labels. Consequently, any unlabeled data point $x_u$ can be seen as a random sample from the combined set of $X_\textit{l}$ and $X_\textit{u}$. This label estimation strategy encourages consistency across different augmentations\citep{67}. The model’s prediction $y_u$ (softmax probabilities) for the augmented copies of each data point $x_u$ is then averaged as follows:
\begin{equation}
\centering
y_{u} =\frac{1}{k} \sum_{a=1}^{k}f(x_{u,a} ;\theta )
\end{equation}

Subsequently, the estimated labels are generated for all unlabeled and augmented data points. The combined labeled and unlabeled datasets are sampled, and all augmented data are used to calculate $L_u$, which is the $L_2$ loss between the predicted softmax probabilities and the estimated labels, as it is less prone to errors from incorrect predictions\citep{68,69}.

In the second stage (SSL-SS), the concept of self-supervised learning is employed to handle semantic segmentation tasks by formulating a pretext task using unlabeled data, such as predicting spatial context, to generate target outputs without supervision\citep{34}. In SSLD, the domain-specific augmentations have been applied to each unlabeled data point $x_u$, creating \textit{k} separate copies. Then, a transformation \textit{T} is applied to these augmented data points. The transformed copies are passed through the segmentation network to obtain predicted labels, which are then subjected to the inverse transformation $T^{-1}$ to bring the guessed labels back to the original spatial space of the data points. The final estimated label for each data point $x_u$ is calculated by averaging the inverse transformed predictions:
\begin{equation}
\centering
y_{u} =\frac{1}{k} \sum_{a=1}^{k}T^{-1}  (f(T(x_{u,a });\theta ))
\end{equation}

We then combine the supervised loss ($L_s$) and unsupervised losses ($L_u$, $L_{ss}$) to form the overall objective function. Random samples from the labeled data $X_l$, denoted as ($X_{l1}$, $Y_{l1}$), are used to compute the supervised loss $L_s$. The supervised loss function is defined as:
\begin{equation}
\centering
L_{s} =L_{CEL} (\lambda _{1} +\lambda _{2}L_{BAL} )+\lambda _{3}L_{DL}+\lambda _{4}L_{SL} 
\end{equation}
where $L_{CEL}$, $L_{BAL}$, $L_{DL}$, and $L_{SL}$ represent cross-entropy loss, boundary-aware loss\citep{35}, dice loss, and surface loss\citep{36}, respectively. This loss schedule prioritizes DL during the initial iterations until convergence, after which $L_{SL}$ is introduced to penalize stray regions\citep{38}. The loss $L_{ss}$ is the $L_2$ loss calculated between the SSLD model’s prediction for the augmented data and the guessed label in SSL-SS. The overall loss function is:
\begin{equation}
\centering
L =L_{S} +\lambda _{u}L_{u}  +\lambda _{ss}L_{ss}   
\end{equation}
where $\lambda _{u}$ and $\lambda _{ss}$ are the respective weighting factors.

\subsubsection{Segmentation Network}

\begin{figure}[ht]
	\centering
 	 \includegraphics[width=0.8\textwidth]{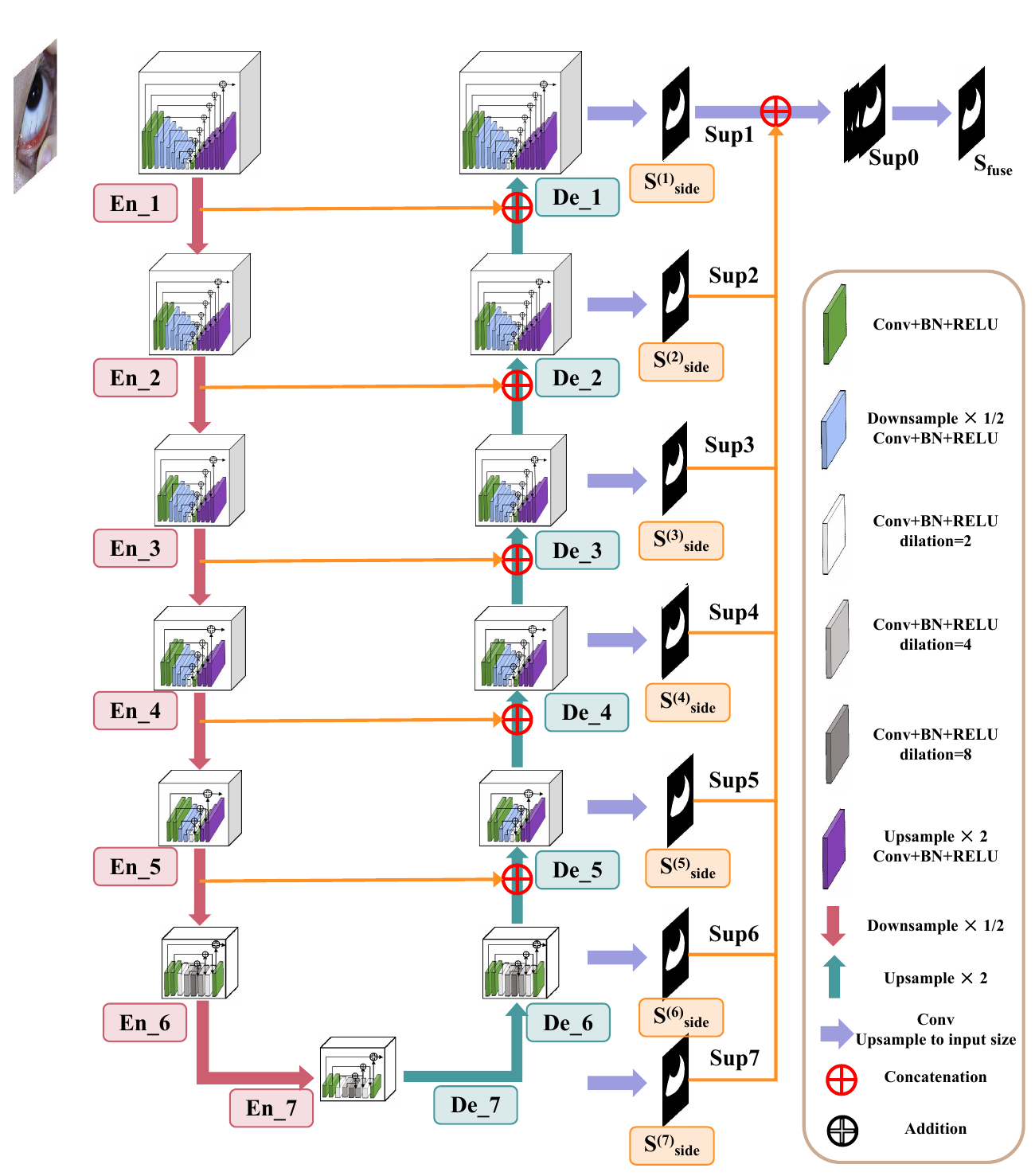}
 	  \caption{Improved U2Net architecture }\label{Figure.3}
 \end{figure}
The segmentation network we utilized is based on an enhanced model of U2Net\citep{33}. U2Net's architecture features a dual-level nested U-shaped structure which efficiently extracts both local and global information from the shallow and deep layers, irrespective of image resolution. This is facilitated by the inclusion of pooling operations in the RSU blocks, which increase the architectural depth without significantly raising the computational cost. Such a design allows for training the network from scratch, eliminating the need for a pre-existing image classification backbone. U2Net has shown exceptional performance across six public salient object detection datasets, outperforming 20 other leading methods in both qualitative and quantitative assessments\citep{33}.

U2Net comprises three primary components: a six-level encoder, a five-level decoder, and a saliency graph fusion model integrated with both the decoder and the final level of the encoder. To enhance image feature extraction, we expanded the encoder to seven levels and the decoder to six levels, introducing an advanced saliency graph fusion module as illustrated in Fig.\ref{Figure.3}. This enhanced network preserves the U-Net style encoder-decoder architecture, utilizing Residual U-blocks (RSU) at each stage. The initial encoder stages, En\_1 through En\_5, employ RSU-8, RSU-7, RSU-6, RSU-5, and RSU-4, respectively, with the number indicating the RSU's height $L$, set based on the spatial resolution of the input feature maps. In the deeper stages, En\_6 and En\_7, where feature map resolution is lower, we use RSU-4F, a variant featuring dilated convolutions to prevent context loss by maintaining input resolution.

The decoding process mirrors the encoder structure, particularly at stage De\_6, where RSU-4F is similarly employed. Each decoding stage integrates upsampled feature maps from the previous decoder stage with feature maps from the corresponding encoder stage. The Saliency Graph Fusion Module then produces salient probability maps. Unlike U2Net, our model initially generates seven saliency probability maps: \(S_{side}^{(7)}\), \(S_{side}^{(6)}\), \(S_{side}^{(5)}\), \(S_{side}^{(4)}\), \(S_{side}^{(3)}\), \(S_{side}^{(2)}\), and \(S_{side}^{(1)}\) from En\_7, De\_6, De\_5, De\_4, De\_3, De\_2, and De\_1, respectively, through a \(3\times3\) convolutional layer, omitting the Sigmoid activation. The output saliency maps are upsampled to the input image size and fused via a cascading operation. The fused result undergoes a \(1\times1\) convolution to produce the final saliency probability map, \(S_{fuse}\).

\section{Experiments and Results}

\subsection{Dataset and Evaluation metrics}
\begin{figure}[tbh]
\begin{center}
\begin{tabular}{ccc}
    \begin{minipage}{0.31\textwidth}
    \includegraphics[width=\textwidth]{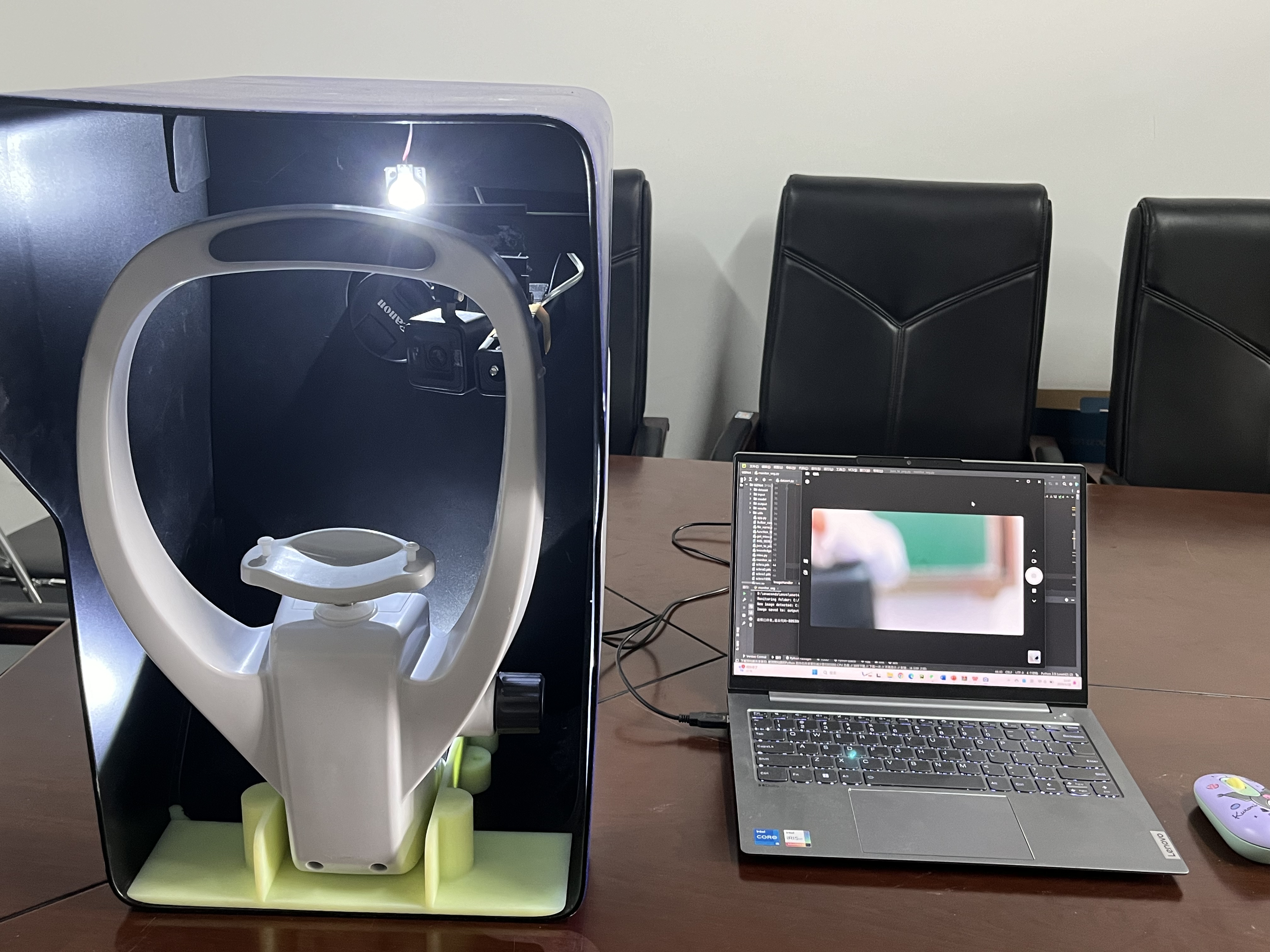}  
    \centerline{(a) Front view}
    \end{minipage} &
    \begin{minipage}{0.31\textwidth}
    \includegraphics[width=\textwidth]{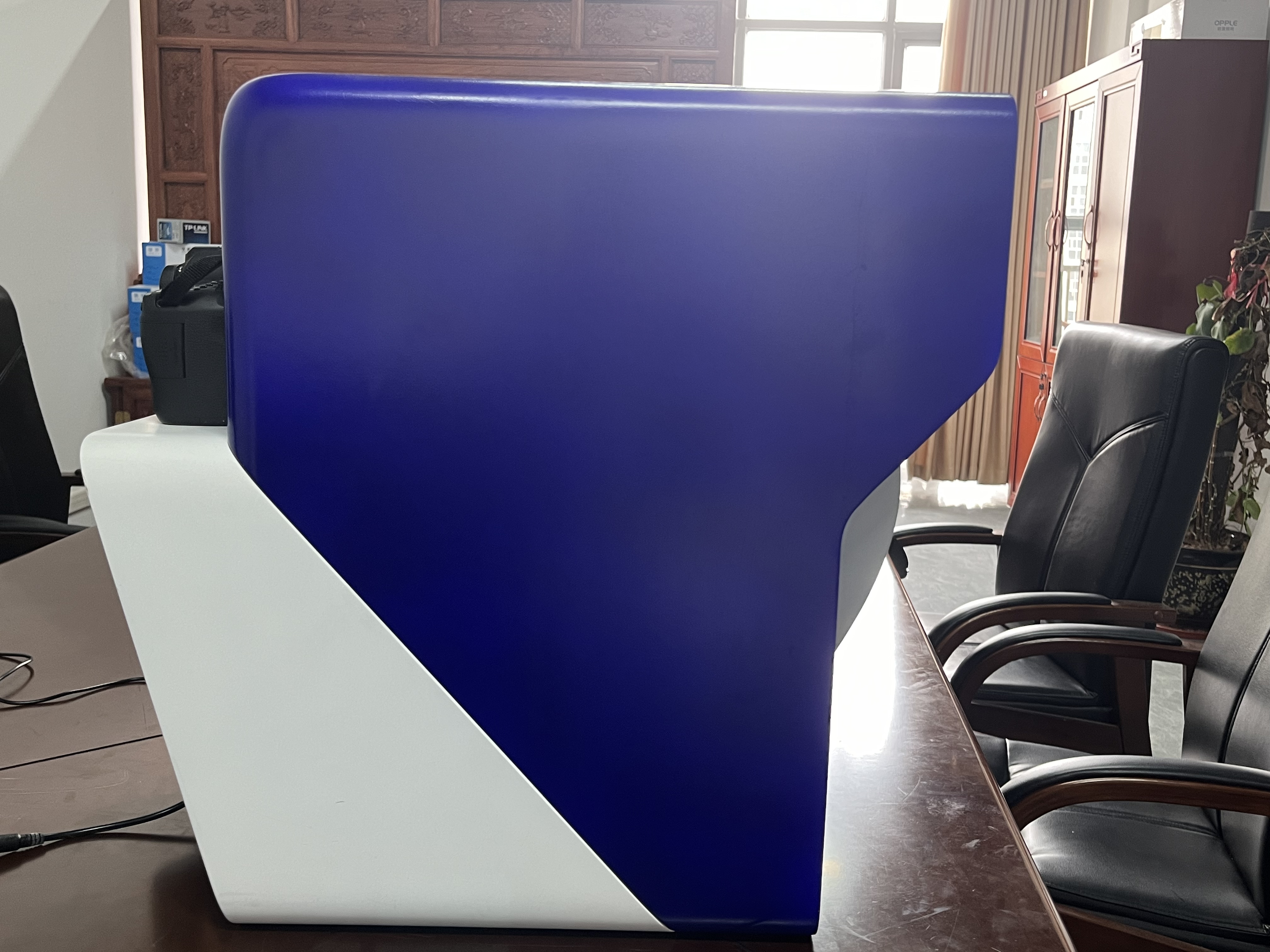}  
    \centerline{(b) Side view}
    \end{minipage} &
    \begin{minipage}{0.31\textwidth}
    \includegraphics[width=\textwidth]{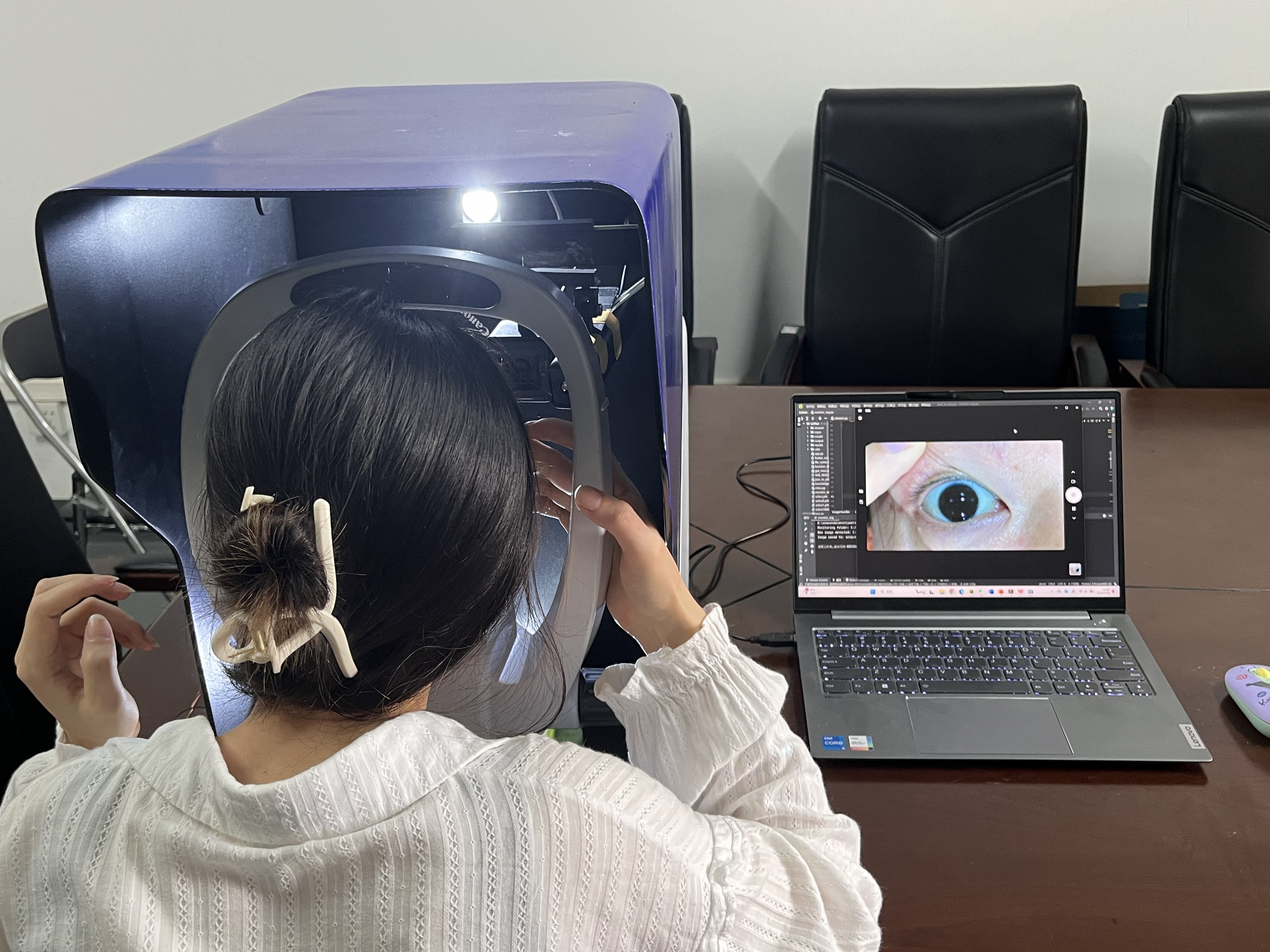}  
    \centerline{(c) Shooting view}
    \end{minipage} \\
\end{tabular}
\end{center}
\caption 
{ \label{Figure.4}
Eye diagnostic instrument} 
\end{figure}

The dataset utilized in this paer was collected using an eye diagnosis instrument developed by our laboratory, as shown in Fig.\ref{Figure.4}. All participants were recruited from the Hainan Provincial Hospital of Traditional Chinese Medicine. For each participant, images of both the left and right eyes were captured, including views from the upward, downward, leftward, and rightward directions. The data labels were manually annotated under the guidance of a physician. To date, the dataset includes eye images from over 100 patients, totaling approximately 800 images. For the purposes of this experiment, about 1000 images were used, segmented into training, validation, and testing sets with a ratio of 7:2:1, respectively.

\begin{figure}[tbh]
\begin{center}
\begin{tabular}{ccc}
    \begin{minipage}{0.25\textwidth}
    \includegraphics[width=\textwidth]{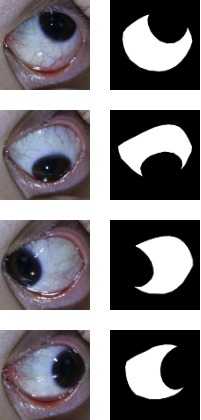}  
    \centerline{(a) Ours}
    \end{minipage} &
    \begin{minipage}{0.25\textwidth}
    \includegraphics[width=\textwidth]{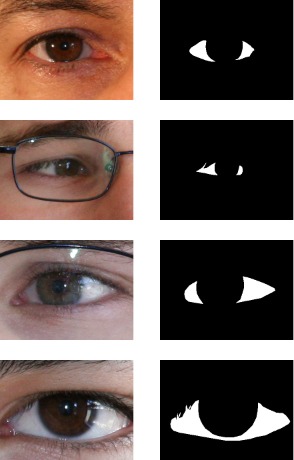}  
    \centerline{(b) UBIRIS.v2}
    \end{minipage} &
    \begin{minipage}{0.25\textwidth}
    \includegraphics[width=\textwidth]{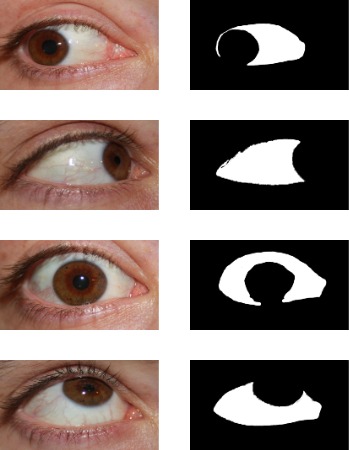}  
    \centerline{(c) SBVPI}
    \end{minipage} \\
\end{tabular}
\end{center}
\caption 
{ \label{Figure.5}
Example images and corresponding sclera segmentation ground truths} 
\end{figure}

We conducted comparative experiments with other networks using two public datasets, UBIRIS.v2 \citep{72} and SBVPI \citep{18}. UBIRIS.v2, initially developed for iris recognition under less constrained conditions, comprises 11,102 images from 261 subjects. From this dataset, a subset of 300 images was manually annotated with sclera masks \citep{19}. We partitioned these 300 images into training, validation, and testing sets using a 4:4:2 ratio, and additionally, we randomly selected 1,000 unlabeled images from this dataset to augment the training set’s unlabeled data. SBVPI, a publicly available database, contains 2,399 high-quality eye images ($3000\times1700$ pixels) from 55 subjects, designed primarily for sclera and periocular recognition research. Each image includes per-pixel sclera annotations. At the time of access, only 1,856 images were available for download from the dataset owner, with 1,836 images having valid sclera annotations. We divided these 1,836 images into training, validation, and testing sets with a 4:4:2 ratio and incorporated the 20 unlabeled images into the training set's unlabeled data pool. Example images and their corresponding sclera segmentation ground truths from UBIRIS.v2, SBVPI, and our datasets are presented in Fig.\ref{Figure.5}. Detailed descriptions of these datasets are summarized in Table~\ref{table1}.

\begin{table}[tbh]
\centering
\caption{Summary of the used datasets. In this case, in the training set part is also divided into labeled and unlabeled data, depending on $X_{l}$. $X_{l}$ is the number of labeled data, and all the rest of the data in the training set are unlabeled data.}\label{table1}%
\begin{tabular}{lllll}
\toprule
Dataset&Resolution& No. of training&No. of validation&No. of testing\\
\midrule
Ours&$5472\times3648$&700&200&100\\
UBIRIS.v2&$400\times300$&1120&120&60\\
SBVPI&$3000\times1700$&754&734&368\\
\bottomrule
\end{tabular}

\end{table}

The performance is assessed using the standard mean Intersection over Union (IoU) score \citep{43}, mean Recall, mean Precision, and F1-score. IoU, a widely utilized metric for semantic segmentation, measures the overlap between the predicted segmentation and the label by dividing the intersection of both by their union. This metric ranges from 0 to 1 (0-100\%), where 0 signifies no overlap and 1 denotes perfect alignment. Recall quantifies the proportion of accurately identified voxels that are part of the actual region of interest, reflecting the model's capability to segment relevant areas effectively. The F1 Score, a statistical measure, evaluates the accuracy of a binary classification model by considering both precision and recall. It represents the harmonic mean of the model's precision and recall, with values ranging from 0 to 1. These metrics are computed as follows:
\begin{equation}
\centering
IoU=\frac{\left | R_{gt}\cap R_{pred}   \right | }{\left | R_{gt}\cup R_{pred}   \right | } =\frac{TP}{TP+FP+FN} 
\end{equation}

\begin{equation}
\centering
Recall=\frac{R_{gt}\cap R_{pred} }{R_{gt}} =\frac{TP}{TP+FN}   
\end{equation}

\begin{equation}
\centering
Precision=PPV=\frac{R_{gt}\cap R_{pred} }{R_{pred}} =\frac{TP}{TP+FP}   
\end{equation}

\begin{equation}
\centering
F_{1} =2\cdot \frac{Precision\cdot Recall}{Precision+Recall}    
\end{equation}
where $R_{gt}$ is the result of the segmentation of ground true, and $R_{pred}$ is the result of the predicted segmentation.

\subsection{Implementation Details}
The models were trained using the Adam optimizer \citep{42} with a learning rate of 0.001 and a batch size comprising four images (two labeled and two unlabeled) across 100 epochs. To expedite computation, all images were resized to 256 $\times$ 256 pixels. In our experiments, the hyperparameters $\lambda _{1}$, $\lambda _{2}$, $\lambda _{3}$ (set as $1-\alpha$), and $\lambda _{4}$ (set as $\alpha$) were defined, where $\alpha$ equals the epoch number divided by 100 for epochs less than 100, and 0 thereafter. The unsupervised loss hyperparameters, $\lambda _{u}$ and $\lambda _{ss}$, were implemented as linearly increasing weights with slopes of 0.02 and 0.002 per epoch, respectively. Initially, the loss scheduling scheme prioritized supervised loss; after several epochs, the emphasis shifted towards unsupervised loss. For domain-specific augmentations, we adjusted the contrast and brightness of the images, selected random values between 0.8 and 1.2 in increments of 0.05 for Gamma correction, and applied random clip parameters and grid sizes for CLAHE set at (1.0, 1.2, 1.5, 1.5, 1.5, 2.0) and (2, 4, 8, 8, 8, 16) respectively. The \textit{T} transform applied random rotations between $-5^{\circ}$ and $5^{\circ}$ and translations between -20 and 20 pixels, with probabilities of 50

\subsection{Experimental Results}
\begin{figure}[tbh]
\begin{center}
\begin{tabular}{cc}
    \begin{minipage}{0.45\textwidth}
    \includegraphics[width=\textwidth]{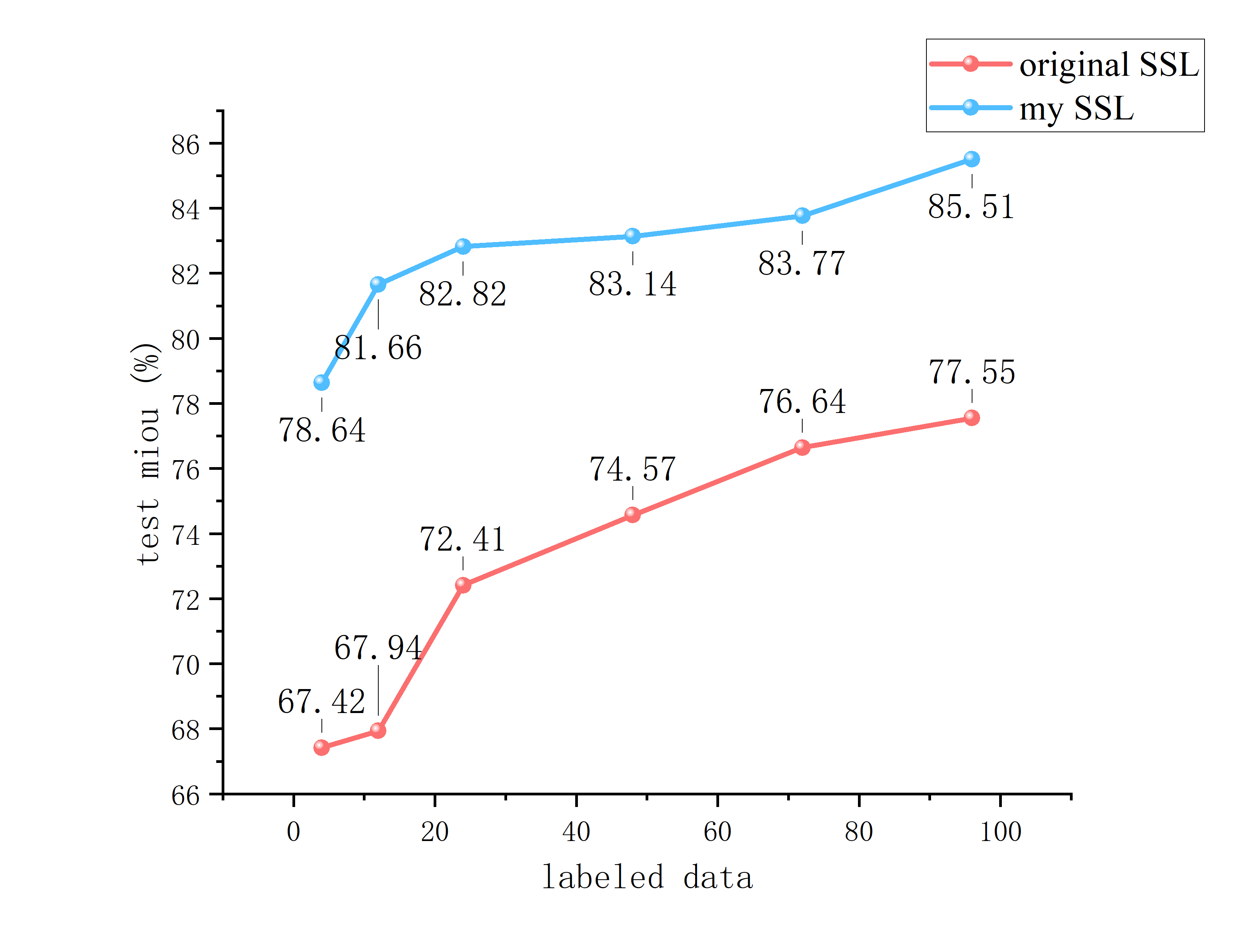}  
    \centerline{(a) mIoU}
    \end{minipage} &
    \begin{minipage}{0.45\textwidth}
    \includegraphics[width=\textwidth]{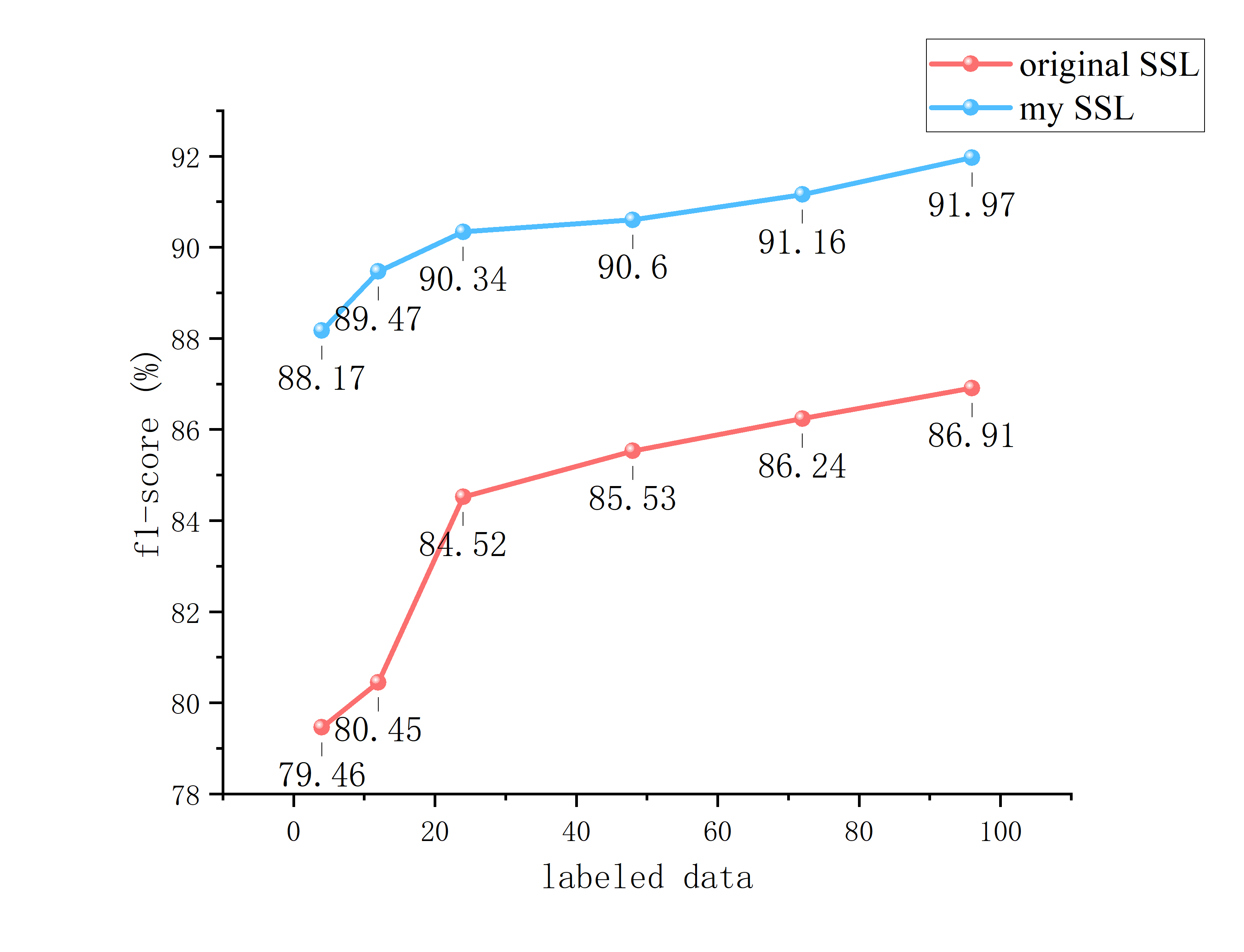}  
    \centerline{(b) F1-score}
    \end{minipage} \\
\end{tabular}
\end{center}
\caption 
{ \label{Figure.6}
The results of the evaluation metrics of the two methods} 
\end{figure}

\paragraph{Evaluation of our semi-supervised method} To verify the effectiveness of our method, we firstly compared our method to the original SSL on our dataset and used the same segmentation network as theirs, RITnet~\citep{38}. Fig.\ref{Figure.6} illustrates the experimental results comparing our semi-supervised learning (SSL) method with the original SSL method, utilizing the same segmentation network but varying the quantity of labeled data. During the experiments, we observed that the original SSL method remained underfitted, as evidenced by the training and validation mean Intersection over Union (mIoU) scores. Notably, our method achieved a better fit within the same 100 epochs, resulting in significantly higher test evaluation metrics compared to the original method. When the number of labeled examples, $X_{l}$, was only 4, our method's test mIoU already exceeded that of the original method at $X_{l} = 96$. This rapid convergence during the training period, coupled with superior performance in fewer epochs, demonstrates the effectiveness of our semi-supervised approach, particularly with limited labeled data 

\begin{figure*}[tbh]
\begin{center}
\begin{tabular}{ccc}
    \begin{minipage}{0.3\textwidth}
    \includegraphics[width=\textwidth]{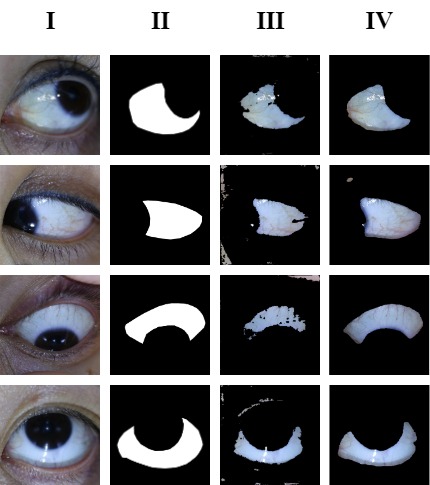}  
    \centerline{(a) 4 labeled images}
    \end{minipage} &
    \begin{minipage}{0.3\textwidth}
    \includegraphics[width=\textwidth]{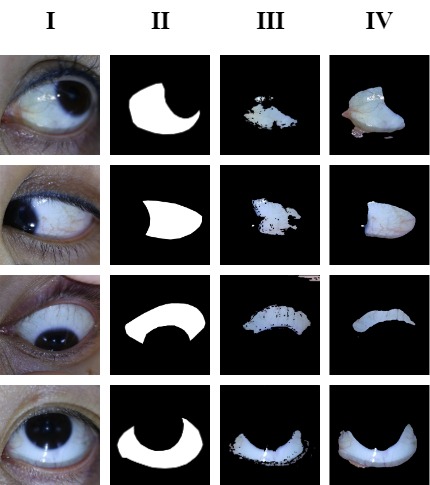}  
    \centerline{(a) 12 labeled images}
    \end{minipage} &
    \begin{minipage}{0.3\textwidth}
    \includegraphics[width=\textwidth]{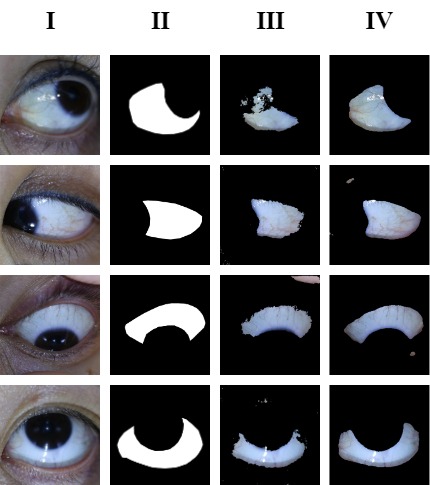}  
    \centerline{(a) 24 labeled images}
    \end{minipage} \\
    \begin{minipage}{0.3\textwidth}
    \includegraphics[width=\textwidth]{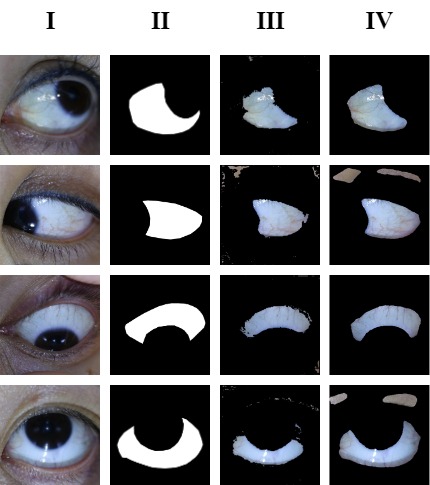}  
    \centerline{(a) 48 labeled images}
    \end{minipage} &
    \begin{minipage}{0.3\textwidth}
    \includegraphics[width=\textwidth]{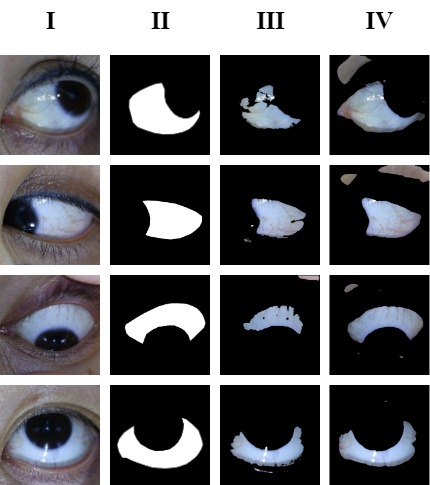}  
    \centerline{(a) 72 labeled images}
    \end{minipage} &
    \begin{minipage}{0.3\textwidth}
    \includegraphics[width=\textwidth]{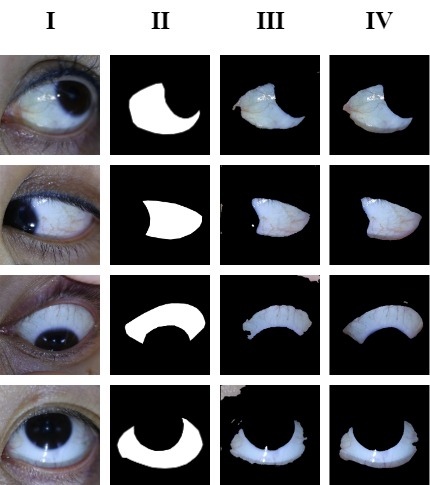}  
    \centerline{(a) 96 labeled images}
    \end{minipage} \\
\end{tabular}
\end{center}
	\caption{Some segmentation results of different SSL method using RITnet as segmentation networks.}
	\label{Figure.7}
\end{figure*}

Fig.\ref{Figure.7} presents segmented results from two methods. Each panel includes four columns: Column \uppercase\expandafter{\romannumeral1} displays the original eye images from our eye diagnosis dataset, Column \uppercase\expandafter{\romannumeral2} shows the ground truths, Column \uppercase\expandafter{\romannumeral3} presents the segmentation results of the original SSL method using RITnet as the segmentation network, and the fourth column depicts the segmentation results of our SSL framework using RITnet. A comparison of the final segmented results with the ground truths reveals that, although our SSL method significantly outperforms the original, neither method effectively segmented the sclera. Consequently, we ultimately chose not to use RITnet as our segmentation network, opting instead for our enhanced U2Net.

\paragraph{Evaluation of different segmentation networks}
We subsequently adopted this improved network for our semi-supervised model. To demonstrate our model's superiority, we tested various segmentation networks within our SSL framework, using different quantities of labeled data, $X_{l}$, as in the earlier experiment. We employed RITnet\citep{38}, UNet\citep{35}, and U2Net\citep{33} for these tests. Table \ref{table2} presents a summary of mIoU, Recall, Precision, and F-measure for these models. The results indicate that using U2Net and our proposed method achieves optimal performance across a range of $X_{l}$ sizes. Notably, our method slightly surpasses the performance of our SSL framework that utilizes U2Net as the segmentation network, according to the evaluation metrics.

\begin{table}[tbh]
\caption{Our semi-supervised method uses four different models as segmentation networks. Experimental results are shown for different $X_{l}$. The bold values represent the best performance in different experimental sets.}\label{table2}

\begin{tabular*}{\linewidth}{@{\extracolsep\fill}cccccc}
\toprule
$X_{l}$ & Segmentation Network & mIoU \% & Recall \% & Precision \% & F1 \% \\
\midrule  
\multirow{4}*{4} & RITnet\citep{38} & 78.64 & 83.13 & 93.86 & 88.17 \\
& UNet\citep{35} & 85.53 & 91.17 & 92.89 & 92.02 \\
& U2Net\citep{33} & 86.45 & \textbf{92.32} & 92.80 & 92.56 \\
& Proposed Method & \textbf{87.94} & 91.38 & \textbf{95.77} & \textbf{93.52} \\
\hline
\multirow{4}*{12} & RITnet\citep{38} & 81.66 & 89.16 & 89.78 & 89.47 \\
& UNet\citep{35} & 86.24 & \textbf{93.28} & 91.55 & 92.41 \\
& U2Net\citep{33} & 86.78 & 92.52 & 92.90 & 92.71 \\
& Proposed Method & \textbf{88.43} & 92.87 & \textbf{94.58} & \textbf{93.72} \\
\hline
\multirow{4}*{24} & RITnet\citep{38} & 82.82 & 92.07 & 88.68 & 90.34 \\
& UNet\citep{35} & 86.34 & 91.10 & 93.91 & 92.48 \\
& U2Net\citep{33} & 87.06 & 92.44 & 93.31 & 92.87 \\
& Proposed Method & \textbf{88.59} & \textbf{92.58} & \textbf{95.12} & \textbf{93.83} \\
\hline
\multirow{4}*{48} & RITnet\citep{38} & 83.14 & 92.69 & 88.60 & 90.60 \\
& UNet\citep{35} & 86.56 & 91.53 & 93.69 & 92.60 \\
& U2Net\citep{33} & 87.34 & 92.27 & 93.86 & 93.06 \\
& Proposed Method & \textbf{88.69} & \textbf{93.18} & \textbf{94.56} & \textbf{93.86} \\
\hline
\multirow{4}*{72} & RITnet\citep{38} & 83.77 & \textbf{94.20} & 88.31 & 91.61 \\
& UNet\citep{35} & 87.46 & 92.53 & 93.73 & 93.13 \\
& U2Net\citep{33} & 87.78 & 91.54 & \textbf{95.33} & 93.40 \\
& Proposed Method & \textbf{89.17} & 93.24 & 95.09 & \textbf{94.16} \\
\hline
\multirow{4}*{96} & RITnet\citep{38} & 85.51 & 93.00 & 90.96 & 91.97 \\
& UNet\citep{35} & 87.65 & 92.39 & 94.12 & 93.25 \\
& U2Net\citep{33} & 88.31 & 91.11 & \textbf{96.69} & 93.82 \\
& Proposed Method & \textbf{89.90} & \textbf{93.65} & 95.52 & \textbf{94.58} \\
\bottomrule
\end{tabular*}
\end{table}

\begin{figure}[tbh]
\begin{center}
\begin{tabular}{cc}
    \begin{minipage}{0.45\textwidth}
    \includegraphics[width=\textwidth]{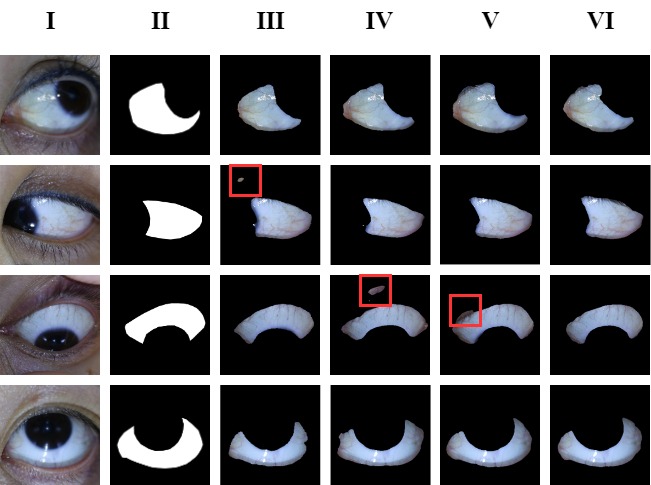}  
    \centerline{(a) 4 labeled images}
    \end{minipage} &
    \begin{minipage}{0.45\textwidth}
    \includegraphics[width=\textwidth]{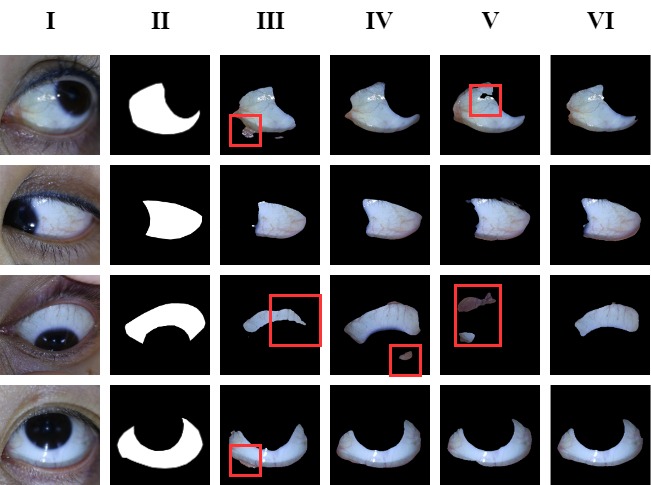}  
    \centerline{(b) 12 labeled images}
    \end{minipage} \\
    \begin{minipage}{0.45\textwidth}
    \includegraphics[width=\textwidth]{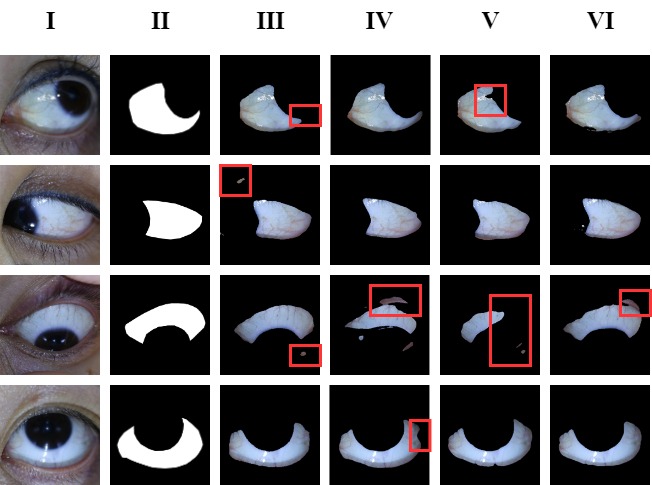}  
    \centerline{(c) 24 labeled images}
    \end{minipage} &
    \begin{minipage}{0.45\textwidth}
    \includegraphics[width=\textwidth]{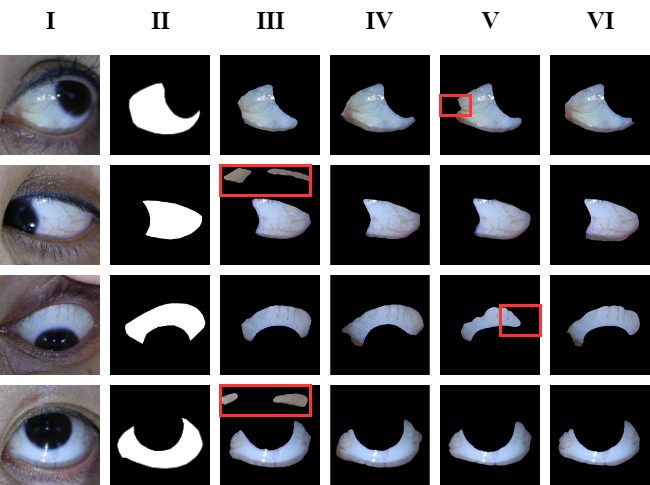}  
    
    \centerline{(d) 48 labeled images}
    \end{minipage} \\
    \begin{minipage}{0.45\textwidth}
    \includegraphics[width=\textwidth]{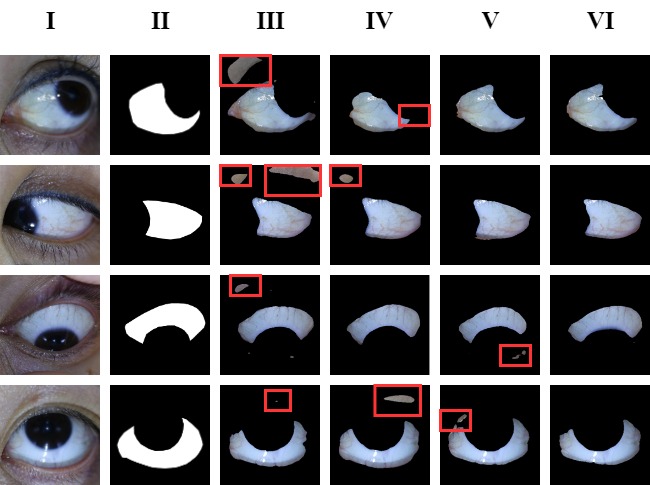}  
    \centerline{(e) 72 labeled images}
    \end{minipage} &
    \begin{minipage}{0.45\textwidth}
    \includegraphics[width=\textwidth]{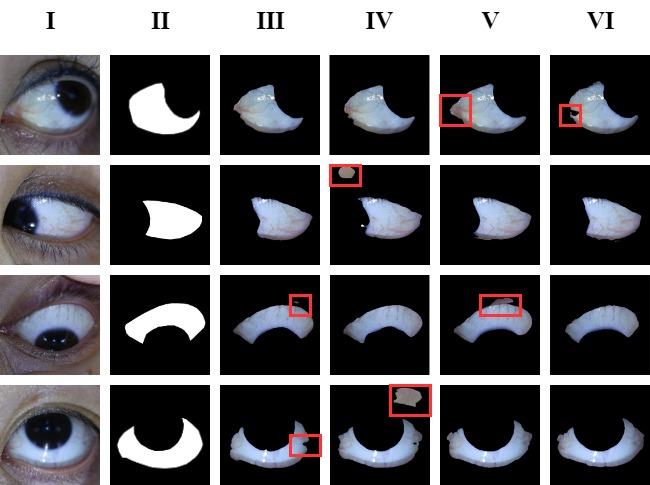}  
    \centerline{(f) 96 labeled images}
    \end{minipage} \\
\end{tabular}
\end{center}
\caption 
{ \label{Figure.8}
Some segmentation results of our SSL method using different segmentation networks} 
\end{figure}

Figure~\ref{Figure.8} displays segmentation results from various models. Each panel includes multiple columns: Column \uppercase\expandafter{\romannumeral1} presents the original eye images from our eye diagnosis dataset; Column \uppercase\expandafter{\romannumeral2} shows the ground truths; Columns \uppercase\expandafter{\romannumeral3} to \uppercase\expandafter{\romannumeral5} depict the segmentation results of our SSL framework using RITnet, UNet, and U2Net as segmentation networks, respectively; the final column illustrates the results from our proposed method. Areas within red boxes highlight regions where the sclera was incorrectly predicted or missed by the corresponding method. Our method demonstrated superior efficacy in segmenting the sclera, aligning with conclusions drawn from the experimental data. It yielded the most region of interest (ROI) coverage, with smoother and more precise edge handling compared to the other methods. Additionally, even with as few as four labeled examples, the segmentation results were still accurate, underscoring the effectiveness of our semi-supervised approach with limited labeled data. These findings also suggest that our enhanced U2Net is more suitable than the other three networks for our SSL framework's segmentation tasks, further validating the efficacy of our novel semi-supervised segmentation method in sclera segmentation.

\paragraph{Evaluation on public datasets}\label{subsubsec5}
To demonstrate that our proposed model is not only valid on our own dataset, we also experimented on two public datasets, UBIRIS.v2 and SBVPI, and the Table \ref{table3} is the list of some models that performed scleral segmentation tasks on the two datasets.

\begin{table}[tbh]
\caption{Some methods perform well on the two datasets for the scleral segmentation task, and the bold values represent the best performance in different datasets, - represents none.}\label{table3}
\centering
\begin{tabular*}{\textwidth}{@{\extracolsep\fill}cccccc}
\toprule
Dataset & Method & mIoU \% & Recall \% & Precision \% & F1 \% \\
\midrule
\multirow{7}{*}{UBIRIS.v2} & ScleraSegNet (CBAM)\citep{28} & 84.21 & 91.42 & 91.59 & 91.26 \\
                           & ScleraSegNet (SSBC)\citep{28} & 84.33 & 92.24 & 90.93 & 91.37 \\
                           & Sclera-Net\citep{27} & - & 89.18 & 92.55 & 90.56 \\
                           & FCN\citep{19} & - & 87.31 & 88.45 & 87.48 \\
                           & UNet\citep{35} & 83.12 & 90.51 & 91.81 & 90.89 \\
                           & SegNet\citep{21} & - & 72.48 & 87.52 & 77.82 \\
                           & Sclera-TransFuse\citep{73} & \textbf{89.69} & \textbf{94.47} & \textbf{94.45} & \textbf{94.53} \\
\midrule
\multirow{6}{*}{SBVPI}     & ScleraSegNet (CBAM)\citep{28} & 91.33 & 95.39 & 95.62 & 95.41 \\
                           & ScleraSegNet (SSBC)\citep{28} & 91.55 & 95.86 & 95.39 & 95.53 \\
                           & Sclera-Net\citep{27} & - & 91.80 & \textbf{98.17} & 94.40 \\
                           & UNet\citep{35} & 91.18 & 95.18 & 95.66 & 95.32 \\
                           & SegNet\citep{21} & 86.17 & \textbf{96.87} & 88.76 & 92.53 \\
                           & Sclera-TransFuse\citep{73} & \textbf{93.59} & 96.82 & 96.59 & \textbf{96.66} \\
\bottomrule
\end{tabular*}
\end{table}

\begin{table}[tbh]
\caption{The evaluation results of our model on the two public datasets, $X_{l}$ represents the number of labeled images in the train set.}\label{table4}
\centering
\begin{tabular*}{\linewidth}{@{\extracolsep\fill}cccccc}
\toprule
Dataset & $X_{l}$ & mIoU \% & Recall \% & Precision \% & F1 \% \\
\midrule
\multirow{6}{*}{UBIRIS.v2} & 4   & 74.21 & 76.82 & 93.83 & 84.48 \\
                            & 12  & 78.13 & 80.99 & 94.16 & 87.08 \\
                            & 24  & 82.22 & 85.63 & 94.13 & 89.68 \\
                            & 48  & 82.93 & 85.41 & 95.80 & 90.31 \\
                            & 72  & 84.60 & 87.93 & 94.77 & 91.22 \\
                            & 96  & 84.95 & 88.06 & 95.15 & 91.47 \\
\midrule
\multirow{7}{*}{SBVPI}      & 4   & 86.56 & 92.34 & 92.69 & 92.51 \\
                            & 12  & 87.07 & 92.33 & 93.35 & 92.84 \\
                            & 24  & 87.32 & 91.45 & 94.67 & 93.03 \\
                            & 48  & 87.92 & 91.74 & 95.13 & 93.40 \\
                            & 72  & 88.21 & 92.35 & 94.79 & 93.55 \\
                            & 96  & 88.50 & 91.42 & 96.35 & 93.82 \\
                            & 500 & 91.77 & 95.47 & 95.75 & 95.61 \\
\bottomrule
\end{tabular*}
\end{table}
The experimental results of our proposed model on two public datasets are presented in Table~\ref{table4}. The data indicate that for the UBIRIS.v2 dataset, as the number of labeled examples ($X_{l}$) increases, the test evaluation metrics improve. Specifically, when $X_{l}$ reaches 72, the mean Intersection over Union (mIoU) attains 84.60\%, surpassing the performance of most models listed in Table~\ref{table3}, which were trained with 120 labeled examples.

\begin{figure}[tbh]
	\centering
 	 \includegraphics[width=\textwidth]{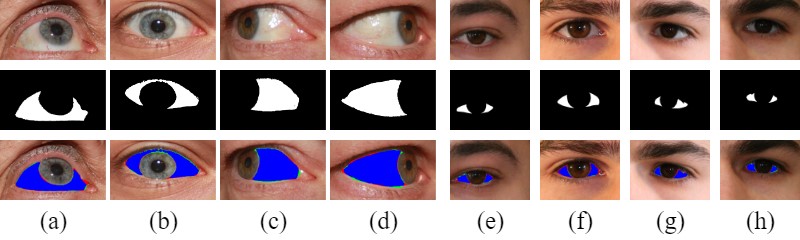}
 	  \caption{Sample images of the segmentation results of using proposed SSL method on UBIRIS.v2 and SBVPI datasets}\label{Figure.9}
 \end{figure}
Given that the SBVPI dataset is larger than ours, we increased the number of epochs to 200 for this section, allowing the model to learn more features. In addition to replicating experiments with the same number of labeled examples ($X_{l}$) as before, we also increased $X_{l}$. Notably, at $X_{l} = 500$, the mean Intersection over Union (mIoU) reached 91.77\%, surpassing the performance of ScleraSegNet\citep{28}, which utilized 734 labeled examples in their training set. Furthermore, even with only four labeled examples, the mIoU of our method exceeded that of SegNet listed in Table~\ref{table3}. Using $X_{l} = 72$ as an example, Fig.\ref{Figure.9} presents segmented images from both datasets. The first row displays the original eye images, with (a-d) from SBVPI and (e-h) from UBIRIS.v2. The second row shows the masks, and the third row illustrates the segmentation results, where blue, green, and red regions denote true positive, false positive, and false negative pixels, respectively.

\section{Conclusions}\label{sec5}
In conclusion, this paper proposes a novel sclera segmentation framework based on semi-supervised learning to address the significant challenges in real-world applications, particularly the scarcity of labeled samples which hampers the performance of deep learning-based methods due to the high cost of acquiring quality sclera data. Our framework utilizes an enhanced U2Net as the segmentation backbone, coupled with effective data augmentation techniques, allowing it to achieve excellent performance with substantially fewer labeled samples. Moreover, we have developed a new eye diagnostic dataset containing approximately 800 images from over 100 patients, each manually annotated by experts, to enhance the evaluation pipeline. Extensive experiments on our dataset and two additional public datasets confirm the effectiveness and superiority of our proposed framework. In future work, we aim to optimize the trade-offs between performance and efficiency while safeguarding patient privacy.

\subsubsection*{Acknowledgments}
This study was supported by Hainan Province Science and Technology Special Fund (ZDYF2022SHFZ304) and the Statistical Information Center of Hainan Provincial Health Commission unveiled the leading project(C2024010117).

\bibliography{iclr2025_conference}

\begin{thebibliography}{75}
\providecommand{\natexlab}[1]{#1}
\providecommand{\url}[1]{\texttt{#1}}
\expandafter\ifx\csname urlstyle\endcsname\relax
  \providecommand{\doi}[1]{doi: #1}\else
  \providecommand{\doi}{doi: \begingroup \urlstyle{rm}\Url}\fi

\bibitem[Chen et~al.(2023{\natexlab{a}})Chen, Ren, Wang, Huang, and Xue]{chen2023interpretable}
Shengchao Chen, Sufen Ren, Guanjun Wang, Mengxing Huang, and Chenyang Xue.
\newblock Interpretable cnn-multilevel attention transformer for rapid recognition of pneumonia from chest x-ray images.
\newblock \emph{IEEE Journal of Biomedical and Health Informatics}, 28\penalty0 (2):\penalty0 753--764, 2023{\natexlab{a}}.

\bibitem[Ren et~al.(2024{\natexlab{a}})Ren, Hu, Chen, and Wang]{ren2024federated}
Sufen Ren, Yule Hu, Shengchao Chen, and Guanjun Wang.
\newblock Federated distillation for medical image classification: Towards trustworthy computer-aided diagnosis.
\newblock \emph{arXiv preprint arXiv:2407.02261}, 2024{\natexlab{a}}.

\bibitem[Peng et~al.(2023)Peng, Chen, Niu, Wang, Yao, Wang, Wang, Tang, Wang, Huang, et~al.]{peng2023diffusion}
Hong Peng, Shengchao Chen, Ning Niu, Jiahao Wang, Qiaoyi Yao, Lu~Wang, Yixuan Wang, Jurui Tang, Guanjun Wang, Mengxing Huang, et~al.
\newblock Diffusion model in semi-supervised scleral segmentation.
\newblock In \emph{2023 IEEE 6th International conference on pattern recognition and artificial intelligence (PRAI)}, pages 376--382. IEEE, 2023.

\bibitem[Wang et~al.(2024{\natexlab{a}})Wang, Peng, Chen, and Ren]{wang2024ensemble}
Jiahao Wang, Hong Peng, Shengchao Chen, and Sufen Ren.
\newblock Ensemble learning for retinal disease recognition under limited resources.
\newblock \emph{Medical \& Biological Engineering \& Computing}, pages 1--14, 2024{\natexlab{a}}.

\bibitem[Shi et~al.(2023)Shi, Chen, Yu, Ren, Wang, and Xue]{shi2023recognition}
Jianying Shi, Shengchao Chen, Benguo Yu, Yi~Ren, Guanjun Wang, and Chenyang Xue.
\newblock Recognition system for diagnosing pneumonia and bronchitis using children’s breathing sounds based on transfer learning.
\newblock \emph{Intelligent Automation \& Soft Computing}, 37\penalty0 (3), 2023.

\bibitem[Alshakree et~al.(2024)Alshakree, Akbas, and Rahebi]{alshakree2024human}
Firas Alshakree, Ayhan Akbas, and Javad Rahebi.
\newblock Human identification using palm print images based on deep learning methods and gray wolf optimization algorithm.
\newblock \emph{Signal, Image and Video Processing}, 18\penalty0 (1):\penalty0 961--973, 2024.

\bibitem[Acharya et~al.(2015)Acharya, Tan, Koh, Sudarshan, Yeo, Too, Chua, Ng, and Tong]{acharya2015automated}
U~Rajendra Acharya, Jen~Hong Tan, Joel~EW Koh, Vidya~K Sudarshan, Sharon Yeo, Cheah~Loon Too, Chua~Kuang Chua, EYK Ng, and Louis Tong.
\newblock Automated diagnosis of dry eye using infrared thermography images.
\newblock \emph{Infrared Physics \& Technology}, 71:\penalty0 263--271, 2015.

\bibitem[Noori et~al.(2016)Noori, Bilonick, and Eller]{noori2016scleral}
Jila Noori, Richard~A Bilonick, and Andrew~W Eller.
\newblock Scleral buckle surgery for primary retinal detachment without posterior vitreous detachment.
\newblock \emph{Retina}, 36\penalty0 (11):\penalty0 2066--2071, 2016.

\bibitem[Mookiah et~al.(2012)Mookiah, Acharya, Lim, Petznick, and Suri]{mookiah2012data}
Muthu Rama~Krishnan Mookiah, U~Rajendra Acharya, Choo~Min Lim, Andrea Petznick, and Jasjit~S Suri.
\newblock Data mining technique for automated diagnosis of glaucoma using higher order spectra and wavelet energy features.
\newblock \emph{Knowledge-Based Systems}, 33:\penalty0 73--82, 2012.

\bibitem[Zhao and Kumar(2017)]{zhao2017towards}
Zijing Zhao and Ajay Kumar.
\newblock Towards more accurate iris recognition using deeply learned spatially corresponding features.
\newblock In \emph{Proceedings of the IEEE international conference on computer vision}, pages 3809--3818, 2017.

\bibitem[Naqvi and Loh(2019)]{27}
Rizwan~Ali Naqvi and Woong-Kee Loh.
\newblock Sclera-net: Accurate sclera segmentation in various sensor images based on residual encoder and decoder network.
\newblock \emph{IEEE Access}, 7:\penalty0 98208--98227, 2019.

\bibitem[Das et~al.(2021)Das, De~Ghosh, and Chattopadhyay]{29}
Sumanta Das, Ishita De~Ghosh, and Abir Chattopadhyay.
\newblock An efficient deep sclera recognition framework with novel sclera segmentation, vessel extraction and gaze detection.
\newblock \emph{Signal Processing: Image Communication}, 97:\penalty0 116349, 2021.

\bibitem[Li et~al.(2023)Li, Wang, Zhao, He, Wang, and Sun]{73}
Haiqing Li, Caiyong Wang, Guangzhe Zhao, Zhaofeng He, Yunlong Wang, and Zhenan Sun.
\newblock Sclera-transfuse: Fusing swin transformer and cnn for accurate sclera segmentation.
\newblock In \emph{2023 IEEE International Joint Conference on Biometrics (IJCB)}, pages 1--8. IEEE, 2023.

\bibitem[Chen et~al.(2023{\natexlab{b}})Chen, Long, Jiang, Liu, and Zhang]{chen2023foundation}
Shengchao Chen, Guodong Long, Jing Jiang, Dikai Liu, and Chengqi Zhang.
\newblock Foundation models for weather and climate data understanding: A comprehensive survey.
\newblock \emph{arXiv preprint arXiv:2312.03014}, 2023{\natexlab{b}}.

\bibitem[Chen et~al.(2024{\natexlab{a}})Chen, Shu, Zhao, Wang, Ren, and Yang]{chen2024free}
Shengchao Chen, Ting Shu, Huan Zhao, Jiahao Wang, Sufen Ren, and Lina Yang.
\newblock Free lunch for federated remote sensing target fine-grained classification: A parameter-efficient framework.
\newblock \emph{Knowledge-Based Systems}, 294:\penalty0 111694, 2024{\natexlab{a}}.

\bibitem[Radu et~al.(2015)Radu, Ferryman, and Wild]{17}
Petru Radu, James Ferryman, and Peter Wild.
\newblock A robust sclera segmentation algorithm.
\newblock In \emph{2015 IEEE 7th International Conference on Biometrics Theory, Applications and Systems (BTAS)}, pages 1--6. IEEE, 2015.

\bibitem[Rot et~al.(2018)Rot, Emer{\v{s}}i{\v{c}}, Struc, and Peer]{18}
Peter Rot, {\v{Z}}iga Emer{\v{s}}i{\v{c}}, Vitomir Struc, and Peter Peer.
\newblock Deep multi-class eye segmentation for ocular biometrics.
\newblock In \emph{2018 IEEE international work conference on bioinspired intelligence (IWOBI)}, pages 1--8. IEEE, 2018.

\bibitem[Ronneberger et~al.(2015)Ronneberger, Fischer, and Brox]{35}
Olaf Ronneberger, Philipp Fischer, and Thomas Brox.
\newblock U-net: Convolutional networks for biomedical image segmentation.
\newblock In \emph{Medical Image Computing and Computer-Assisted Intervention--MICCAI 2015: 18th International Conference, Munich, Germany, October 5-9, 2015, Proceedings, Part III 18}, pages 234--241. Springer, 2015.

\bibitem[Chen et~al.(2023{\natexlab{c}})Chen, Shu, Zhao, Zhong, and Chen]{chen2023tempee}
Shengchao Chen, Ting Shu, Huan Zhao, Guo Zhong, and Xunlai Chen.
\newblock Tempee: Temporal-spatial parallel transformer for radar echo extrapolation beyond auto-regression.
\newblock \emph{IEEE Transactions on Geoscience and Remote Sensing}, 2023{\natexlab{c}}.

\bibitem[Fuhl et~al.(2016)Fuhl, Santini, Kasneci, and Kasneci]{fuhl2016pupilnet}
Wolfgang Fuhl, Thiago Santini, Gjergji Kasneci, and Enkelejda Kasneci.
\newblock Pupilnet: Convolutional neural networks for robust pupil detection.
\newblock \emph{arXiv preprint arXiv:1601.04902}, 2016.

\bibitem[{\'S}wirski et~al.(2012){\'S}wirski, Bulling, and Dodgson]{swirski2012robust}
Lech {\'S}wirski, Andreas Bulling, and Neil Dodgson.
\newblock Robust real-time pupil tracking in highly off-axis images.
\newblock In \emph{Proceedings of the symposium on eye tracking research and applications}, pages 173--176, 2012.

\bibitem[Chen et~al.(2022{\natexlab{a}})Chen, Shu, Zhao, Wan, Huang, and Li]{chen2022dynamic}
Shengchao Chen, Ting Shu, Huan Zhao, Qilin Wan, Jincan Huang, and Cailing Li.
\newblock Dynamic multiscale fusion generative adversarial network for radar image extrapolation.
\newblock \emph{IEEE Transactions on Geoscience and Remote Sensing}, 60:\penalty0 1--11, 2022{\natexlab{a}}.

\bibitem[Chen et~al.(2023{\natexlab{d}})Chen, Long, Shen, and Jiang]{chen2023prompt}
Shengchao Chen, Guodong Long, Tao Shen, and Jing Jiang.
\newblock Prompt federated learning for weather forecasting: Toward foundation models on meteorological data.
\newblock \emph{arXiv preprint arXiv:2301.09152}, 2023{\natexlab{d}}.

\bibitem[Ren et~al.(2024{\natexlab{b}})Ren, Chen, Wang, Xu, Hou, Huang, Liu, and Wang]{ren2024distributed}
Sufen Ren, Shengchao Chen, Jiahao Wang, Haoyang Xu, Xuan Hou, Mengxing Huang, Jianxun Liu, and Guanjun Wang.
\newblock A distributed photonic crystal fiber reverse design framework based on multi-source knowledge fusion.
\newblock \emph{Optical Fiber Technology}, 84:\penalty0 103718, 2024{\natexlab{b}}.

\bibitem[Chen et~al.(2024{\natexlab{b}})Chen, Long, Jiang, and Zhang]{chen2024personalized}
Shengchao Chen, Guodong Long, Jing Jiang, and Chengqi Zhang.
\newblock Personalized adapter for large meteorology model on devices: Towards weather foundation models.
\newblock \emph{arXiv preprint arXiv:2405.20348}, 2024{\natexlab{b}}.

\bibitem[Zhou et~al.(2003)Zhou, Bousquet, Lal, Weston, and Sch{\"o}lkopf]{zhou2003learning}
Dengyong Zhou, Olivier Bousquet, Thomas Lal, Jason Weston, and Bernhard Sch{\"o}lkopf.
\newblock Learning with local and global consistency.
\newblock \emph{Advances in neural information processing systems}, 16, 2003.

\bibitem[Tarvainen and Valpola(2017)]{tarvainen2017mean}
Antti Tarvainen and Harri Valpola.
\newblock Mean teachers are better role models: Weight-averaged consistency targets improve semi-supervised deep learning results.
\newblock \emph{Advances in neural information processing systems}, 30, 2017.

\bibitem[Laine and Aila(2016)]{68}
Samuli Laine and Timo Aila.
\newblock Temporal ensembling for semi-supervised learning.
\newblock \emph{arXiv preprint arXiv:1610.02242}, 2016.

\bibitem[Gyawali et~al.(2019)Gyawali, Li, Ghimire, and Wang]{gyawali2019semi}
Prashnna~Kumar Gyawali, Zhiyuan Li, Sandesh Ghimire, and Linwei Wang.
\newblock Semi-supervised learning by disentangling and self-ensembling over stochastic latent space.
\newblock In \emph{Medical Image Computing and Computer Assisted Intervention--MICCAI 2019: 22nd International Conference, Shenzhen, China, October 13--17, 2019, Proceedings, Part VI 22}, pages 766--774. Springer, 2019.

\bibitem[Ouali et~al.(2020)Ouali, Hudelot, and Tami]{ouali2020semi}
Yassine Ouali, C{\'e}line Hudelot, and Myriam Tami.
\newblock Semi-supervised semantic segmentation with cross-consistency training.
\newblock In \emph{Proceedings of the IEEE/CVF conference on computer vision and pattern recognition}, pages 12674--12684, 2020.

\bibitem[Chaudhary et~al.(2021)Chaudhary, Gyawali, Wang, and Pelz]{32}
Aayush~Kumar Chaudhary, Prashnna~K Gyawali, Linwei Wang, and Jeff~B Pelz.
\newblock Semi-supervised learning for eye image segmentation.
\newblock In \emph{ACM Symposium on Eye Tracking Research and Applications}, pages 1--7, 2021.

\bibitem[Wang et~al.(2024{\natexlab{b}})Wang, Peng, Chen, and Ren]{wang2024less}
Jiahao Wang, Hong Peng, Shengchao Chen, and Sufen Ren.
\newblock Less is more: Ensemble learning for retinal disease recognition under limited resources.
\newblock \emph{arXiv preprint arXiv:2402.09747}, 2024{\natexlab{b}}.

\bibitem[Chen et~al.(2023{\natexlab{e}})Chen, Shu, Zhao, and Tang]{chen2023mask}
Shengchao Chen, Ting Shu, Huan Zhao, and Yuan~Yan Tang.
\newblock Mask-cnn-transformer for real-time multi-label weather recognition.
\newblock \emph{Knowledge-Based Systems}, 278:\penalty0 110881, 2023{\natexlab{e}}.

\bibitem[Derakhshani et~al.(2006)Derakhshani, Ross, and Crihalmeanu]{3}
Reza Derakhshani, Arun Ross, and Simona Crihalmeanu.
\newblock A new biometric modality based on conjunctival vasculature.
\newblock In \emph{Proceedings of artificial neural networks in engineering}, pages 1--8, 2006.

\bibitem[Derakhshani and Ross(2007)]{4}
Reza Derakhshani and Arun Ross.
\newblock A texture-based neural network classifier for biometric identification using ocular surface vasculature.
\newblock In \emph{2007 International Joint Conference on Neural Networks}, pages 2982--2987. IEEE, 2007.

\bibitem[Crihalmeanu et~al.(2009)Crihalmeanu, Ross, and Derakhshani]{5}
Simona Crihalmeanu, Arun Ross, and Reza Derakhshani.
\newblock Enhancement and registration schemes for matching conjunctival vasculature.
\newblock In \emph{Advances in Biometrics: Third International Conference, ICB 2009, Alghero, Italy, June 2-5, 2009. Proceedings 3}, pages 1240--1249. Springer, 2009.

\bibitem[Thomas et~al.(2010)Thomas, Du, and Zhou]{6}
N~Luke Thomas, Yingzi Du, and Zhi Zhou.
\newblock A new approach for sclera vein recognition.
\newblock In \emph{Mobile Multimedia/Image Processing, Security, and Applications 2010}, volume 7708, pages 38--47. SPIE, 2010.

\bibitem[Oh and Toh(2012)]{7}
Kangrok Oh and Kar-Ann Toh.
\newblock Extracting sclera features for cancelable identity verification.
\newblock In \emph{2012 5th IAPR International Conference on Biometrics (ICB)}, pages 245--250. IEEE, 2012.

\bibitem[Lin et~al.(2013)Lin, Du, Zhou, and Thomas]{8}
Yong Lin, Eliza~Yingzi Du, Zhi Zhou, and N~Luke Thomas.
\newblock An efficient parallel approach for sclera vein recognition.
\newblock \emph{IEEE Transactions on Information Forensics and Security}, 9\penalty0 (2):\penalty0 147--157, 2013.

\bibitem[Zhou et~al.(2010)Zhou, Du, and Thomas]{9}
Zhi Zhou, Eliza~Y Du, and N~Luke Thomas.
\newblock A comprehensive sciera image quality measure.
\newblock In \emph{2010 11th International Conference on Control Automation Robotics \& Vision}, pages 638--643. IEEE, 2010.

\bibitem[Zhou et~al.(2012)Zhou, Du, Belcher, Thomas, and Delp]{10}
Zhi Zhou, Eliza~Yingzi Du, Craig Belcher, N~Luke Thomas, and Edward~J Delp.
\newblock Quality fusion based multimodal eye recognition.
\newblock In \emph{2012 IEEE International Conference on Systems, Man, and Cybernetics (SMC)}, pages 1297--1302. IEEE, 2012.

\bibitem[Khosravi and Safabakhsh(2008)]{11}
Mohammad~Hossein Khosravi and Reza Safabakhsh.
\newblock Human eye sclera detection and tracking using a modified time-adaptive self-organizing map.
\newblock \emph{Pattern Recognition}, 41\penalty0 (8):\penalty0 2571--2593, 2008.

\bibitem[Crihalmeanu and Ross(2012)]{12}
Simona Crihalmeanu and Arun Ross.
\newblock Multispectral scleral patterns for ocular biometric recognition.
\newblock \emph{Pattern Recognition Letters}, 33\penalty0 (14):\penalty0 1860--1869, 2012.

\bibitem[Das et~al.(2013)Das, Pal, Ballester, and Blumenstein]{13}
Abhijit Das, Umapada Pal, Miguel Angel~Ferrer Ballester, and Michael Blumenstein.
\newblock Sclera recognition using dense-sift.
\newblock In \emph{2013 13th International Conference on Intellient Systems Design and Applications}, pages 74--79. IEEE, 2013.

\bibitem[Das et~al.(2014)Das, Pal, Ballester, and Blumenstein]{14}
Abhijit Das, Umapada Pal, Miguel Angel~Ferrer Ballester, and Michael Blumenstein.
\newblock A new efficient and adaptive sclera recognition system.
\newblock In \emph{2014 IEEE Symposium on Computational Intelligence in Biometrics and Identity Management (CIBIM)}, pages 1--8. IEEE, 2014.

\bibitem[Delna et~al.(2016)Delna, Sneha, and Aneesh]{15}
KV~Delna, KA~Sneha, and RP~Aneesh.
\newblock Sclera vein identification in real time using single board computer.
\newblock In \emph{2016 International Conference on Next Generation Intelligent Systems (ICNGIS)}, pages 1--5. IEEE, 2016.

\bibitem[Alkassar et~al.(2016)Alkassar, Woo, Dlay, and Chambers]{16}
Sinan Alkassar, Wai~Lok Woo, Satnam~Singh Dlay, and Jonathon~A Chambers.
\newblock Enhanced segmentation and complex-sclera features for human recognition with unconstrained visible-wavelength imaging.
\newblock In \emph{2016 International Conference on Biometrics (ICB)}, pages 1--8. IEEE, 2016.

\bibitem[Lucio et~al.(2018)Lucio, Laroca, Severo, Britto, and Menotti]{19}
Diego~R Lucio, Rayson Laroca, Evair Severo, Alceu~S Britto, and David Menotti.
\newblock Fully convolutional networks and generative adversarial networks applied to sclera segmentation.
\newblock In \emph{2018 IEEE 9th International Conference on Biometrics Theory, Applications and Systems (BTAS)}, pages 1--7. IEEE, 2018.

\bibitem[Rot et~al.(2020)Rot, Vitek, Grm, Emer{\v{s}}i{\v{c}}, Peer, and {\v{S}}truc]{20}
Peter Rot, Matej Vitek, Klemen Grm, {\v{Z}}iga Emer{\v{s}}i{\v{c}}, Peter Peer, and Vitomir {\v{S}}truc.
\newblock Deep sclera segmentation and recognition.
\newblock \emph{Handbook of vascular biometrics}, pages 395--432, 2020.

\bibitem[Badrinarayanan et~al.(2017)Badrinarayanan, Kendall, and Cipolla]{21}
Vijay Badrinarayanan, Alex Kendall, and Roberto Cipolla.
\newblock Segnet: A deep convolutional encoder-decoder architecture for image segmentation.
\newblock \emph{IEEE transactions on pattern analysis and machine intelligence}, 39\penalty0 (12):\penalty0 2481--2495, 2017.

\bibitem[Long et~al.(2015)Long, Shelhamer, and Darrell]{23}
Jonathan Long, Evan Shelhamer, and Trevor Darrell.
\newblock Fully convolutional networks for semantic segmentation.
\newblock In \emph{Proceedings of the IEEE conference on computer vision and pattern recognition}, pages 3431--3440, 2015.

\bibitem[Isola et~al.(2017)Isola, Zhu, Zhou, and Efros]{25}
Phillip Isola, Jun-Yan Zhu, Tinghui Zhou, and Alexei~A Efros.
\newblock Image-to-image translation with conditional adversarial networks.
\newblock In \emph{Proceedings of the IEEE conference on computer vision and pattern recognition}, pages 1125--1134, 2017.

\bibitem[Teichmann et~al.(2018)Teichmann, Weber, Zoellner, Cipolla, and Urtasun]{24}
Marvin Teichmann, Michael Weber, Marius Zoellner, Roberto Cipolla, and Raquel Urtasun.
\newblock Multinet: Real-time joint semantic reasoning for autonomous driving.
\newblock In \emph{2018 IEEE intelligent vehicles symposium (IV)}, pages 1013--1020. IEEE, 2018.

\bibitem[Boutros et~al.(2019)Boutros, Damer, Kirchbuchner, and Kuijper]{26}
Fadi Boutros, Naser Damer, Florian Kirchbuchner, and Arjan Kuijper.
\newblock Eye-mms: Miniature multi-scale segmentation network of key eye-regions in embedded applications.
\newblock In \emph{Proceedings of the IEEE/CVF International Conference on Computer Vision Workshops}, pages 0--0, 2019.

\bibitem[Wang et~al.(2019)Wang, Wang, Liu, He, He, and Sun]{28}
Caiyong Wang, Yunlong Wang, Yunfan Liu, Zhaofeng He, Ran He, and Zhenan Sun.
\newblock Sclerasegnet: An attention assisted u-net model for accurate sclera segmentation.
\newblock \emph{IEEE Transactions on Biometrics, Behavior, and Identity Science}, 2\penalty0 (1):\penalty0 40--54, 2019.

\bibitem[Wang et~al.(2023)Wang, Li, Ma, Zhao, and He]{30}
Caiyong Wang, Haiqing Li, Wenhui Ma, Guangzhe Zhao, and Zhaofeng He.
\newblock Metascleraseg: an effective meta-learning framework for generalized sclera segmentation.
\newblock \emph{Neural Computing and Applications}, 35\penalty0 (29):\penalty0 21797--21826, 2023.

\bibitem[Wang et~al.(2022)Wang, Ren, Li, Chen, and Yu]{wang2022semi}
Gao Wang, Sufen Ren, Shuna Li, Shengchao Chen, and Benguo Yu.
\newblock Semi-supervised deep learning model for efficient computation of optical properties of suspended-core fibers.
\newblock \emph{Sensors}, 22\penalty0 (18):\penalty0 6751, 2022.

\bibitem[Ren et~al.(2023)Ren, Chen, Xu, Hou, Yang, Wang, and Shen]{ren2023unsupervised}
Sufen Ren, Shengchao Chen, Haoyang Xu, Xuan Hou, Qian Yang, Guanjun Wang, and Chong Shen.
\newblock Unsupervised representation learning-based spectrum reconstruction for demodulation of fabry--p{\'e}rot interferometer sensor.
\newblock \emph{IEEE Sensors Journal}, 23\penalty0 (12):\penalty0 13810--13816, 2023.

\bibitem[Jocher et~al.(2022)Jocher, Chaurasia, Stoken, Borovec, Kwon, Michael, Fang, Wong, Yifu, Montes, et~al.]{74}
Glenn Jocher, Ayush Chaurasia, Alex Stoken, Jirka Borovec, Yonghye Kwon, Kalen Michael, Jiacong Fang, Colin Wong, Zeng Yifu, Diego Montes, et~al.
\newblock ultralytics/yolov5: v6. 2-yolov5 classification models, apple m1, reproducibility, clearml and deci. ai integrations.
\newblock \emph{Zenodo}, 2022.

\bibitem[Xu et~al.(2023)Xu, Chen, Ren, Hou, Wang, and Shen]{xu2023dual}
Haoyang Xu, Shengchao Chen, Sufen Ren, Xuan Hou, Guanjun Wang, and Chong Shen.
\newblock Dual-parameter demodulation of fbg-fpi cascade sensors via sparse samples: A deep learning-based perspective.
\newblock \emph{IEEE Sensors Journal}, 2023.

\bibitem[Chen et~al.(2022{\natexlab{b}})Chen, Yao, Ren, Yang, Yang, Yuan, Wang, and Huang]{chen2022fabry}
Shengchao Chen, Feifan Yao, Sufen Ren, Jianli Yang, Qian Yang, Shuyu Yuan, Guanjun Wang, and Mengxing Huang.
\newblock Fabry-perot interferometric sensor demodulation system utilizing multi-peak wavelength tracking and neural network algorithm.
\newblock \emph{Optics express}, 30\penalty0 (14):\penalty0 24461--24480, 2022{\natexlab{b}}.

\bibitem[Chen et~al.(2022{\natexlab{c}})Chen, Ren, Wang, Huang, and Xue]{chen2022cmt}
Shengchao Chen, Sufen Ren, Guanjun Wang, Mengxing Huang, and Chenyang Xue.
\newblock Cmt: Interpretable model for rapid recognition pneumonia from chest x-ray images by fusing low complexity multilevel attention mechanism.
\newblock \emph{arXiv preprint arXiv:2210.16584}, 2022{\natexlab{c}}.

\bibitem[Chen et~al.(2022{\natexlab{d}})Chen, Ren, Yang, Yao, Yang, Wang, Wang, and Huang]{chen2022reconstruction}
Shengchao Chen, Sufen Ren, Jianli Yang, Feifan Yao, Qian Yang, Lu~Wang, Guanjun Wang, and Mengxing Huang.
\newblock Reconstruction of fabry-perot interferometric sensor spectrum from extremely sparse sampling points using dense neural network.
\newblock \emph{IEEE Photonics Technology Letters}, 34\penalty0 (24):\penalty0 1337--1340, 2022{\natexlab{d}}.

\bibitem[Pizer et~al.(1987)Pizer, Amburn, Austin, Cromartie, Geselowitz, Greer, ter Haar~Romeny, Zimmerman, and Zuiderveld]{pizer1987adaptive}
Stephen~M Pizer, E~Philip Amburn, John~D Austin, Robert Cromartie, Ari Geselowitz, Trey Greer, Bart ter Haar~Romeny, John~B Zimmerman, and Karel Zuiderveld.
\newblock Adaptive histogram equalization and its variations.
\newblock \emph{Computer vision, graphics, and image processing}, 39\penalty0 (3):\penalty0 355--368, 1987.

\bibitem[Zuiderveld(1994)]{37}
Karel Zuiderveld.
\newblock Contrast limited adaptive histogram equalization.
\newblock \emph{Graphics gems}, pages 474--485, 1994.

\bibitem[Chaudhary et~al.(2019)Chaudhary, Kothari, Acharya, Dangi, Nair, Bailey, Kanan, Diaz, and Pelz]{38}
Aayush~K Chaudhary, Rakshit Kothari, Manoj Acharya, Shusil Dangi, Nitinraj Nair, Reynold Bailey, Christopher Kanan, Gabriel Diaz, and Jeff~B Pelz.
\newblock Ritnet: Real-time semantic segmentation of the eye for gaze tracking.
\newblock In \emph{2019 IEEE/CVF International Conference on Computer Vision Workshop (ICCVW)}, pages 3698--3702. IEEE, 2019.

\bibitem[Guan et~al.(2009)Guan, Jian, Hongda, Zhiguo, and Haibin]{guan2009image}
Xu~Guan, Su~Jian, Pan Hongda, Zhang Zhiguo, and Gong Haibin.
\newblock An image enhancement method based on gamma correction.
\newblock In \emph{2009 Second international symposium on computational intelligence and design}, volume~1, pages 60--63. IEEE, 2009.

\bibitem[Gyawali et~al.(2020)Gyawali, Ghimire, Bajracharya, Li, and Wang]{67}
Prashnna~Kumar Gyawali, Sandesh Ghimire, Pradeep Bajracharya, Zhiyuan Li, and Linwei Wang.
\newblock Semi-supervised medical image classification with global latent mixing.
\newblock In \emph{Medical Image Computing and Computer Assisted Intervention--MICCAI 2020: 23rd International Conference, Lima, Peru, October 4--8, 2020, Proceedings, Part I 23}, pages 604--613. Springer, 2020.

\bibitem[Berthelot et~al.(2019)Berthelot, Carlini, Goodfellow, Papernot, Oliver, and Raffel]{69}
David Berthelot, Nicholas Carlini, Ian Goodfellow, Nicolas Papernot, Avital Oliver, and Colin~A Raffel.
\newblock Mixmatch: A holistic approach to semi-supervised learning.
\newblock \emph{Advances in neural information processing systems}, 32, 2019.

\bibitem[Kolesnikov et~al.(2019)Kolesnikov, Zhai, and Beyer]{34}
Alexander Kolesnikov, Xiaohua Zhai, and Lucas Beyer.
\newblock Revisiting self-supervised visual representation learning.
\newblock In \emph{Proceedings of the IEEE/CVF conference on computer vision and pattern recognition}, pages 1920--1929, 2019.

\bibitem[Kervadec et~al.(2019)Kervadec, Bouchtiba, Desrosiers, Granger, Dolz, and Ayed]{36}
Hoel Kervadec, Jihene Bouchtiba, Christian Desrosiers, Eric Granger, Jose Dolz, and Ismail~Ben Ayed.
\newblock Boundary loss for highly unbalanced segmentation.
\newblock In \emph{International conference on medical imaging with deep learning}, pages 285--296. PMLR, 2019.

\bibitem[Qin et~al.(2020)Qin, Zhang, Huang, Dehghan, Zaiane, and Jagersand]{33}
Xuebin Qin, Zichen Zhang, Chenyang Huang, Masood Dehghan, Osmar~R Zaiane, and Martin Jagersand.
\newblock U2-net: Going deeper with nested u-structure for salient object detection.
\newblock \emph{Pattern recognition}, 106:\penalty0 107404, 2020.

\bibitem[Proen{\c{c}}a et~al.(2009)Proen{\c{c}}a, Filipe, Santos, Oliveira, and Alexandre]{72}
Hugo Proen{\c{c}}a, Silvio Filipe, Ricardo Santos, Joao Oliveira, and Luis~A Alexandre.
\newblock The ubiris. v2: A database of visible wavelength iris images captured on-the-move and at-a-distance.
\newblock \emph{IEEE Transactions on Pattern Analysis and Machine Intelligence}, 32\penalty0 (8):\penalty0 1529--1535, 2009.

\bibitem[Everingham et~al.(2006)Everingham, Zisserman, Williams, Van~Gool, Allan, Bishop, Chapelle, Dalal, Deselaers, Dork{\'o}, et~al.]{43}
Mark Everingham, Andrew Zisserman, Christopher~KI Williams, Luc Van~Gool, Moray Allan, Christopher~M Bishop, Olivier Chapelle, Navneet Dalal, Thomas Deselaers, Gyuri Dork{\'o}, et~al.
\newblock The 2005 pascal visual object classes challenge.
\newblock In \emph{Machine Learning Challenges. Evaluating Predictive Uncertainty, Visual Object Classification, and Recognising Tectual Entailment: First PASCAL Machine Learning Challenges Workshop, MLCW 2005, Southampton, UK, April 11-13, 2005, Revised Selected Papers}, pages 117--176. Springer, 2006.

\bibitem[Kingma and Ba(2014)]{42}
Diederik~P Kingma and Jimmy Ba.
\newblock Adam: A method for stochastic optimization.
\newblock \emph{arXiv preprint arXiv:1412.6980}, 2014.

\end{thebibliography}
\bibliographystyle{unsrtnat} 

\appendix

\end{document}